%% file: article.tex
\documentclass[12pt,letterpaper]{article}
\usepackage[a4paper, total={7in, 10in}]{geometry}

\usepackage{graphicx}
\usepackage{helvet}
\usepackage{authblk}
\usepackage{hyperref}
\usepackage{amsmath}
\usepackage{amssymb} 
\usepackage{orcidlink} 
\usepackage[super,comma,sort&compress]
   {natbib}

\newcommand{\citeauthorsup}[1]{\citeauthor{#1}\textsuperscript{\cite{#1}}}

\input{helperFunctions}
\usepackage{booktabs}
\usepackage[frozencache,cachedir=_minted-article]{minted}
\usepackage{makecell}
\usepackage{subcaption}
\usepackage{caption}
\usepackage{tablefootnote}
\usepackage{pdfpages}
\usepackage{tabularx}
\usepackage{ragged2e}

\makeatletter
\renewcommand{\maketitle}{\bgroup\setlength{\parindent}{0pt}
\begin{flushleft}
  \textbf{\@title}
  
  \@author
\end{flushleft}\egroup}
\makeatother

\title{mlr3mbo: Bayesian Optimization in R}
\date{}

\author[1,2]{Marc Becker}
\author[1,2]{Lennart Schneider}
\author[1,2]{Martin Binder}
\author[3]{Lars Kotthoff}
\author[1,2,4,5]{Bernd Bischl}

\affil[1]{Department of Statistics, LMU Munich, Munich, Germany}
\affil[2]{Munich Center for Machine Learning (MCML), Munich, Germany}
\affil[3]{University of St Andrews, St Andrews, Scotland}
\affil[4]{Senior author}
\affil[5]{Lead contact}

\affil[*]{Correspondence: bernd.bischl@stat.uni-muenchen.de}

\begin{document}

\maketitle

\section*{SUMMARY}

We present \texttt{mlr3mbo}, a modular toolbox for Bayesian optimization in R.
\texttt{mlr3mbo} supports single- and multi-objective optimization, multi-point proposals, batch and asynchronous parallelization, and robust error handling.
While it can be used for many standard Bayesian optimization variants in applied settings, researchers can also construct custom Bayesian optimization algorithms from its flexible building blocks.
In addition to an introduction to the software, its design principles, and its building blocks, the paper presents two extensive empirical evaluations on the surrogate-based benchmark suite YAHPO Gym.
To identify robust default configurations for both numeric and mixed-hierarchical optimization regimes, and to gain further insights into the respective impacts of individual settings, we run a coordinate descent search over the \texttt{mlr3mbo} configuration space and analyze its results.
Furthermore, we benchmark \texttt{mlr3mbo} against a wide range of established optimizers, including \texttt{HEBO}, \texttt{SMAC3}, \texttt{Ax}, and \texttt{Optuna}, and find that it performs on par with state-of-the-art.

\section*{KEYWORDS}

 machine learning, model-based optimization, Bayesian optimization, black-box optimization, AutoML, R, scientific computing, software

\section{Introduction}
\label{sec:introduction}

Black-box optimization deals with optimizing the parameters of an objective function, system, or process whose internal structure is unknown and for which gradients or similar structural information are unavailable.
Instead, the objective function can only be evaluated at given inputs.
These evaluations are often time-consuming, noisy, sometimes involve multiple criteria, and may involve search spaces with integer and categorical parameters as well as hierarchical dependencies.
Such settings are common in engineering and the natural sciences, for example in the design of production or engineering processes via computer simulations~\citep{forrester2008,sacks1989}, drug discovery~\citep{pyzer-knapp2018}, material design~\citep{zhang2020}, agent design~\citep{chen2018}, robotics, hyperparameter tuning of machine learning (ML) models~\citep{snoek2012}, and configuration of general algorithmic systems\cite{chen2018,hutter2011}.
In these regimes, sample-efficient and robust optimization methods are essential to achieving good performance with limited evaluation budgets.

Model-based optimization (MBO), also known as Bayesian optimization (BO) or sequential design of experiments from the field of design and analysis of computer experiments (DACE)\cite{sacks1989}, has become a standard approach for tackling such black-box optimization problems.
Classical BO methods\cite{jones1998} approximate the unknown objective function with a probabilistic surrogate, typically a Gaussian process (GP)\cite{rasmussen2005}, and use an acquisition function, such as expected improvement (EI), to select promising evaluation points.
This leads to a sequential decision process that balances exploitation of regions with low predicted loss and exploration of regions with high uncertainty.
Early work in this area emerged from the simulation and statistics communities~\cite{mockus1978, jones1998}, and has since been generalized to a wide variety of surrogate models (SM) and acquisition strategies~\cite{brochu2010, shahriari2016}.
More historical context can be found in Chapter~12 of Garnett (2023)\cite{garnett2023}; for an applied perspective on GP surrogates and BO see Gramacy (2020)\cite{gramacy2020}.

In ML, BO has become a central tool for hyperparameter optimization (HPO)\cite{bergstra2011, snoek2012, thornton2013, bischl2023}.
Modern learning systems often expose dozens to hundreds of hyperparameters that may be continuous, integer, and categorical, as well as conditional choices over model architectures and preprocessing steps.
Na\"ive approaches such as grid search or purely random search\cite{bergstra2012} are easy to implement, with the latter often showing surprisingly good performance on HPO problems of only moderate difficulty.
However, they do not scale well to problems of higher dimensionality or situations where local information needs to be exploited to reach search space regions of acceptable performance.
This is especially common in AutoML where complete ML pipelines or ensembles are optimized instead of single models. 
As a result, BO is now a core component of many automated machine learning (AutoML) systems~\cite{feurer2022, kotthoff2017}.

In parallel with methodological advances, several software frameworks for BO and HPO have been developed, including \texttt{SMAC} (sequential model-based algorithm configuration)\cite{hutter2011, lindauer2022}, \texttt{Hyperopt}~\cite{bergstra2011}, \texttt{HEBO} (heteroscedastic and evolutionary Bayesian optimization)\cite{cowen-rivers2022}, \texttt{Ax} (adaptive experimentation platform)\cite{olson2025}, \texttt{BoTorch}~\cite{balandat2020} and \texttt{Optuna}~\cite{akiba2019}.
These frameworks differ in the SMs they support, acquisition functions, support for mixed and hierarchical search spaces, and parallelization capabilities.
All statements about features and capabilities of these frameworks refer to the package versions listed in Table~\ref{tab:software-versions}.
In R, the main BO toolkits are \texttt{DiceOptim}, a GP-based EGO toolbox popular in computer experiments, and \texttt{mlrMBO}\cite{bischl2018}, which is \texttt{mlr3mbo}'s predecessor.
For streamlined hyperparameter tuning, the smaller packages \texttt{rBayesianOptimization}\cite{yan2016} and \texttt{ParBayesianOptimization}\cite{wilson2018} provide lightweight GP-based BO with standard acquisition rules.

\paragraph{Main Contributions}

We introduce \texttt{mlr3mbo}, an extensible R toolbox that provides a robust and modular framework for BO.
The package serves multiple goals and different user groups: 
a) It is a flexible BO toolbox, which enables researchers (and students) to study BO variants and implement their own extensions. This is enabled by a modular design philosophy that uses object-oriented components that can be combined and extended seamlessly. 
b) Practitioners can conveniently run a standard, yet highly configurable BO variant on black-box optimization problems. 
c) Hyperparameter optimization (HPO) experiments can easily be run with \texttt{mlr3mbo} as a BO backend, as it is seamlessly integrated into \texttt{mlr3} and \texttt{mlr3tuning} as an HPO optimizer, again with reasonable and well-tested defaults. 

Furthermore, \texttt{mlr3mbo} offers advanced features such as the use of arbitrary regression models as SMs, the handling of mixed and hierarchical search spaces, parallel multi-point proposals (batch and asynchronous), multi-objective optimization, and robust error handling.
Benchmark results demonstrate that \texttt{mlr3mbo} performs competitively compared to state-of-the-art frameworks. Moreover, it has already been used successfully in real-world applications\cite{halac2024,ado2025}.

\section{Bayesian Optimization}

\label{sec:smbo}

\subsection*{General Principle}

Let $\fx: \X \to \R$ be an arbitrary function with a $d$-dimensional input domain $\X = \X_1 \times \X_2 \times \cdots \times \X_d$ and an output $y$ (see Section~\nameref{sec:multi-objective-optimization} for generalizations).
Each $\X_i$ ($i=1,\ldots,d$) can be either numeric (real-valued or integer) and bounded (i.e., $\X_i = [l_i, u_i] \subset \R$ or $\X_i = \{l_i, l_i+1, \ldots, u_i\} \subset \mathbb{Z}$) or it can be a finite set of $s_i$ categorical values ($\X_i = \left\{ v_{i1}, \ldots, v_{is_i} \right\}$).
The goal is to find the input $\xv^\ast$ with

\begin{displaymath}
\xv^\ast \in \argmin_{\xv\in\X} f(\xv).
\end{displaymath}

In the context of BO, it is typically assumed that $f$ is black-box, usually non-convex, and expensive to evaluate, and the total number of function evaluations is limited by a budget.
At the core of BO is an SM $\fh$, which provides an inexpensive, probabilistic approximation of the expensive black-box function $f$ and which is iteratively updated and refined.
BO is a global optimization algorithm that usually proceeds as follows:

\begin{enumerate}
  \item Sample an initial design from $\X$ and evaluate it with $f$ to obtain\\
  $\mathcal{D}_0 = \{(\xv^{(0, 1)}, y^{(0, 1)}), \ldots, (\xv^{(0, n_\text{init})}, y^{(0, n_\text{init})})\}$ with $y^{(0, j)} = f(\xv^{(0, j)})$.
  $\mathcal{D}_t$ is called the ``archive''. 
  \item For $t = 0, 1, 2, \dots$ repeat:
    \begin{enumerate}
      \item Fit an SM $\fh$ on the current archive $\mathcal{D}_t$ or update it with the newly added points.
      \item Establish an acquisition function $\alpha(\xv)$ on top of the SM, which encodes how attractive design points are for subsequent evaluation, and optimize it to obtain the proposed point(s) $\{\xv^{(t+1, 1)}, \ldots, \xv^{(t+1, n_\text{batch})}\}$ for new experiments. 
      \item Evaluate the new point(s) with $f$ and add $\{(\xv^{(t+1, 1)}, y^{(t+1, 1)}), \ldots, (\xv^{(t+1, n_\text{batch})}, y^{(t+1, n_\text{batch})})\}$ to $\mathcal{D}_t$ to obtain $\mathcal{D}_{t+1}$.
      \item Check if the termination criterion is met; if not, return to step (a).
    \end{enumerate}
\end{enumerate}

This effectively substitutes the expensive optimization of $\fx$ in each iteration with the optimization of $\alpha(\xv)$, which inherits the search space structure from $f$, but is now cheap to evaluate. 
We will now briefly discuss the above steps, but also refer the reader to the book by Roman Garnett\cite{garnett2023} and multiple excellent surveys\cite{brochu2010,shahriari2016,frazier2018} for more details.

\subsection*{Initial Design}

\label{sec:initial-design}

The initial design consists of a set of points in the search space $\X$ that are evaluated before fitting the initial SM.
Choosing too few points, or selecting points that do not adequately cover the search space, might result in a degenerate SM and also suboptimal proposed points for the next iteration, whereas a too large initial design consumes a disproportionate amount of the available budget and might slow down sequential learning and optimization of the objective. 
Common design methods are uniform i.i.d.\ random, Latin hypercube (LHS), and Sobol sequence sampling.
LHS divides each dimension into as many intervals as requested points, and samples points to ensure uniform marginal coverage, i.e., no two points share the same bin in any dimension. Additionally, common variants also maximize the minimal distance between points.
Sobol sequences generate quasi-random points that fill the space more uniformly than purely random sampling.
Further details and comparisons can be found in Bossek et al.~(2020)\cite{bossek2020} and Le Riche and Picheny~(2021)\cite{leriche2021}.
Empirical analyses suggest that distance-distributed designs spreading points across a range of pairwise distances can outperform standard space-filling designs for GP surrogates\cite{zhang2021}, and a recent Wasserstein-based characterization shows that the geometry of the initial design has measurable downstream effects on myopic acquisition decisions in BO\cite{candelieri2026}.
The initial design can also be constructed or extended via domain knowledge, if fixed candidates or distributions are available, which often work well on similar problems 
\cite{feurer2015}.

\subsection*{Surrogate Model}

\label{sec:surrogate-model}

The SM $\fh$ is a (usually nonlinear, probabilistic) regression function that approximates the unknown black-box function $f$ using the data observed so far.
It serves as a substitute for the underlying expensive black-box problem that is cheap to evaluate.
A primary factor influencing the choice of SM is the structure of the search space $\X$.
If the search space is purely numerical, GPs\cite{rasmussen2005} are usually the recommended choice\cite{garnett2023}. 
They constitute a local, fully probabilistic regression model that includes posterior uncertainty estimation (see Section~\nameref{sec:acquisition-function} for why this is usually necessary) and can be flexibly adapted via a covariance kernel. 

If the search space includes categorical parameters or hierarchical dependencies between parameters, random forests are a viable alternative, as they can handle such parameters directly\cite{hutter2011,eggensperger2021}.
GPs have also been extended to mixed and categorical search spaces via specialized kernels and encodings\cite{garrido-merchan2020,ru2020,daulton2022}, although this typically requires more modeling care than the direct handling provided by tree-based surrogates.
To enable the use of random forests as surrogates in BO, it is necessary to compute posterior uncertainty estimates, for which several more or less heuristic approaches exist, including jackknife\cite{wager2014}, infinitesimal jackknife\cite{wager2014}, ensemble standard deviation (ESD)\cite{hutter2011}, and the law of total variance (LTV)\cite{hutter2014}.
See Supplement~\nameref{sec:uncertainty-estimation} for more information on uncertainty estimation.

Another important consideration is the computational complexity of the SM.
Vanilla GPs scale with $O(n^3 + p\,n^2)$, where $n$ is the number of data points and $p$ the number of parameters, while random forests scale with $O(T p\,n \log{n})$, where $T$ is the number of trees, with usual feature subsampling.
This makes random forests an attractive choice in large-scale settings with many observations\cite{egele2023}, although a substantial line of recent work has developed scalable GP variants, e.g., sparse inducing-point variational inference\cite{hensman2013}, GPU-accelerated exact inference\cite{wang2019}, or trust-region-based local GPs\cite{eriksson2019}.
Conversely, in the low-data regime typical for expensive black-box optimization, GPs often provide better-calibrated uncertainty estimates and smoother extrapolation than random forests, whose piecewise-constant predictions and heuristic variance estimators can be limiting.

\subsection*{Acquisition Function}

\label{sec:acquisition-function}

The acquisition function quantifies the utility of evaluating a particular point in the search space and guides the optimization process by balancing exploration and exploitation.
It typically combines the posterior mean and posterior standard deviation, both estimated by the SM, into a single criterion.
Regions with low posterior mean likely allow for further local refinement and exploitation, while regions with higher posterior standard deviation correspond to areas of the search space that are less explored and where, therefore, the SM is more uncertain.
The most commonly used acquisition function is expected improvement~(EI)\cite{jones1998, kushner1964, mockus1978}, which estimates the expected gain over the current best observation, lower-bounded at 0:

\begin{displaymath}
\alpha_{\mathrm{EI}}(\xv) = \mathbb{E} \left[ \max \left( f_{\mathrm{min}} - Y(\xv), 0 \right) \right].
\end{displaymath}

Here, $Y(\xv)$ is a random variable following the SM's posterior predictive distribution for a given point $\xv$ and $f_{\mathrm{min}}$ is the best function value found so far.
The lower bound ensures positive gains and leads to positive incorporation of posterior uncertainty and a reward for exploration. 
If $Y(\xv)$ follows a Gaussian distribution, the expected improvement can be calculated in closed form:

\begin{displaymath}
    \alpha_{\mathrm{EI}}(\xv) = (f_{\mathrm{min}} - \mu(\xv)) \Phi\left(\frac{f_{\mathrm{min}} - \mu(\xv)}{\sigma(\xv)}\right) + \sigma(\xv) \phi\left(\frac{f_{\mathrm{min}} - \mu(\xv)}{\sigma(\xv)}\right),
\end{displaymath}

where $\phi$ and $\Phi$ are the probability density function and the cumulative distribution function of the standard normal distribution, respectively, and $\mu(\xv)$ and $\sigma(\xv)$ are the posterior mean and standard deviation of the SM.
A simpler approach to balancing the posterior mean and standard deviation for a point \xv is the lower confidence bound (LCB)\cite{cox1997,srinivas2010}.

\begin{displaymath}
\alpha_{\operatorname{LCB}}(\xv) = \mu{}(\xv) - \lambda \sigma(\xv),
\end{displaymath}

where $\lambda > 0$ is a constant that controls the trade-off between exploitation and exploration.

The LogEI acquisition function\cite{hutter2009,watanabe2024} modifies the standard EI to work correctly with SMs trained on log-transformed target values (see Supplement \nameref{sec:output-transformations} for more details).
This acquisition function is not to be confused with taking the logarithm of the EI for numerical stability\cite{ament2023}.
Many other acquisition functions exist, and various BO variants require custom acquisition functions. 
See the upcoming sections, especially on multi-criteria optimization, and see Table~\ref{tab:acqfunctions}.

\subsection*{Acquisition Function Optimization}

\label{sec:acquisition-optimizer}
Newly proposed points are obtained by optimizing the acquisition function.
Acquisition functions like the EI are usually highly multi-modal\cite{garnett2023}, and gradient availability depends on the choice of SM. 
Common approaches to optimize the acquisition function include simple random search (RS), derivative-free global optimization such as DIRECT (dividing rectangles)\cite{jones1993}, restarted local searches (LS)\cite{hutter2011}, evolutionary algorithms like CMA-ES (Covariance matrix adaptation evolution strategy)\cite{hansen2001}, and restarted gradient-based techniques like L-BFGS-B (limited-memory BFGS with box constraints)\cite{byrd1995}.
Multi-start local optimization methods are frequently employed in combination with purely local searches to address the problem of multi-modality\cite{snoek2012,balandat2020,kim2021}.
A related question is how exactly the acquisition function should be optimized: since the SM is typically misspecified in practice, it has been argued that exact optimization may be unnecessary or even computationally wasteful, and that cheaper, inexact schemes can suffice\cite{ahmed2016}.
The empirical evidence here is mixed, however, and our own benchmark results in Section~\nameref{sec:benchmarks} indicate that investing in stronger acquisition function optimizers with sufficient budget is more often beneficial than not, so the question of how much optimization effort is actually required remains an interesting open problem.

\section{General Structure of \texttt{mlr3mbo}}

\label{sec:mlr3mbo}

\texttt{mlr3mbo} is part of the \texttt{mlr3} ecosystem for ML in R\cite{lang2019}. 
It also builds upon two user-relevant service packages: 
the \texttt{bbotk} (black-box optimization toolkit) for general black-box objectives and optimization, and 
the \texttt{paradox} package, which provides a small domain-specific language for defining parameter spaces.
\texttt{mlr3mbo} can be used in multiple ways to facilitate black-box optimization and hyperparameter tuning.
One option, likely most relevant to researchers and experienced applied data scientists, is to construct custom BO variants using the versatile building blocks provided by \texttt{mlr3mbo}, which we explain in detail.
However, \texttt{mlr3mbo} also provides convenient service functions for running many standard BO variants directly, which we cover later in Section~\nameref{sec:predefined-loop-functions} and, for HPO specifically, in Section~\nameref{sec:hpo}.

\begin{figure}[ht]
  \centering
  \includegraphics[width=0.75\linewidth]{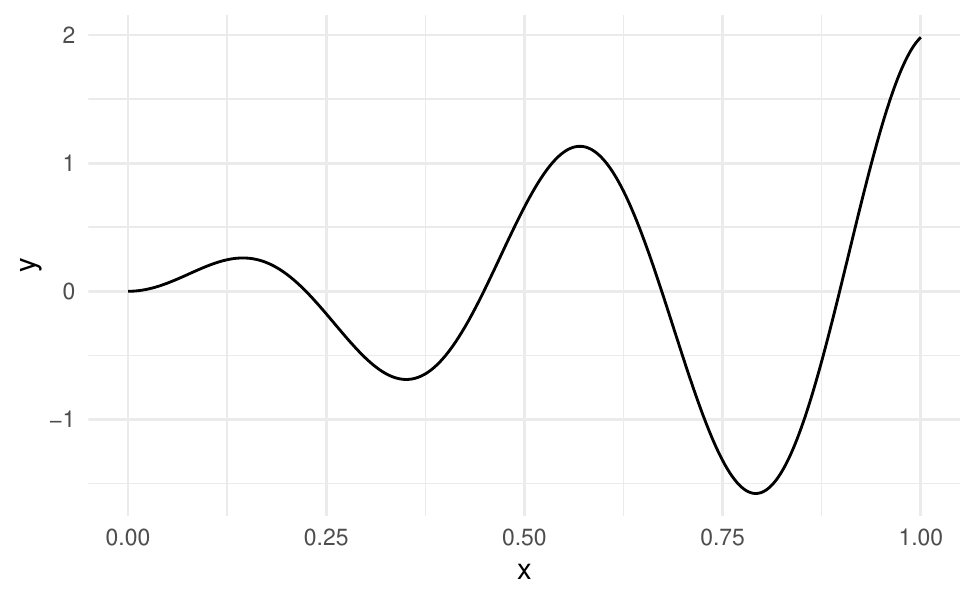}
  \caption{Sinusoidal function used as an example.}
  \label{fig:sinusoidal-function}
\end{figure}

We now demonstrate the functionality of the \texttt{mlr3mbo} R package with the example objective function $f: [0, 1] \rightarrow \mathbb{R}, x \mapsto 2x\,\sin(14x)$ (Figure~\ref{fig:sinusoidal-function}), which has three local minima, one of which is global.

This objective function and its search space can be defined with \texttt{mlr3mbo} with the following code, which also sets up the optimization process.

\begin{minted}{r}
library(mlr3mbo)
library(mlr3learners)
library(bbotk)
sinus_1D = function(xs) list(y = 2 * xs$x * sin(14 * xs$x))

domain = ps(x = p_dbl(lower = 0, upper = 1))
codomain = ps(y = p_dbl(tags = "minimize"))
objective = ObjectiveRFun$new(sinus_1D, domain = domain, codomain = codomain)
instance = oi(objective,
  search_space = domain,
  terminator = trm("evals", n_evals = 20))
\end{minted}

The \texttt{Objective} class represents the target function, with subclasses for different data types and interfaces. 
For example, the \texttt{ObjectiveRFun} class allows users to specify an objective as an R function that takes a list argument (to allow for different data types in each dimension) as input. 
Here, the objective takes a scalar, real-valued $x$ and returns a scalar $y$.
The domain and codomain of the objective are specified using a parameter set from the \texttt{paradox} package.
A parameter set is constructed with the \texttt{ps()} function, which takes named parameters, including their data types and constraints, as arguments (see Table~\ref{tab:param-constructors}).
By tagging a dimension of the codomain with \texttt{minimize} or \texttt{maximize}, we specify the optimization direction.
More details on the \texttt{paradox} package can be found in the HPO and \texttt{paradox} chapters of the \texttt{mlr3} book\cite{becker2024}.
\texttt{mlr3mbo} provides dictionaries for acquisition functions, optimizers, terminators, and loop functions that store predefined R6 objects so users can easily list, search, and access them. 
Values in these dictionaries are typically accessed through sugar functions that retrieve objects from the relevant dictionary; for example, \texttt{acqf("EI")} is a convenience wrapper around \texttt{mlr\_acqfunctions\$get("ei")}.

\begin{table}
  \centering
  \begin{tabular}{lll}
    \toprule
    Constructor     & Description                            & Underlying class  \\
    \midrule
    \texttt{p\_dbl} & Real valued parameter (``double'')     & \texttt{ParamDbl} \\
    \texttt{p\_int} & Integer parameter                      & \texttt{ParamInt} \\
    \texttt{p\_fct} & Discrete valued parameter (``factor'') & \texttt{ParamFct} \\
    \texttt{p\_lgl} & Logical / Boolean parameter            & \texttt{ParamLgl} \\
    \texttt{p\_uty} & Untyped parameter                      & \texttt{ParamUty} \\
    \bottomrule
  \end{tabular}
  \caption{Parameter constructors and their underlying classes.}
  \label{tab:param-constructors}
\end{table}

The \texttt{OptimInstance} class (from \texttt{bbotk}), which is constructed via the \texttt{oi()} sugar function, bundles the objective, the search space, an archive of past evaluations, and termination criteria.
The \texttt{trm("evals", n\_evals = 20)} call creates a \texttt{Terminator} object that terminates the optimization process after 20 evaluations (see Section~\nameref{sec:termination}).

\texttt{OptimizerMbo} is the base class for all MBO variants and holds all the key elements: the BO algorithm loop structure \texttt{loop\_function}, SM \texttt{Surrogate},  acquisition function \texttt{AcqFunction}, and acquisition function optimizer \texttt{AcqOptimizer}.

\subsection*{Initial Design}

Initial designs are usually constructed by calling one of \texttt{paradox}'s generators:

\begin{itemize}
  \item \texttt{generate\_design\_lhs()}: Latin hypercube sampling
  \item \texttt{generate\_design\_random()}: Random sampling
  \item \texttt{generate\_design\_grid()}: Grid sampling
  \item \texttt{generate\_design\_sobol()}: Sobol sequence
\end{itemize}

Here, we construct an initial design of three points using random sampling.

\begin{minted}{r}
sample_domain = ps(x1 = p_dbl(0, 1), x2 = p_dbl(0, 1))
design = generate_design_random(sample_domain, n = 3)$data
\end{minted}

Users may also specify a custom design as a \texttt{data.table}. 
The design can be evaluated using the \texttt{\$eval\_batch()} method of the instance.

\begin{minted}{r}
design = data.table(x = c(0.1, 0.34, 0.65, 1))
instance$eval_batch(design)
instance$archive$data

#        x          y  x_domain           timestamp batch_nr
#    <num>      <num>    <list>              <POSc>    <int>
# 1:  0.10  0.1970899 <list[1]> 2025-06-13 12:14:57        1
# 2:  0.34 -0.6792294 <list[1]> 2025-06-13 12:14:57        1
# 3:  0.65  0.4148279 <list[1]> 2025-06-13 12:14:57        1
# 4:  1.00  1.9812147 <list[1]> 2025-06-13 12:14:57        1
\end{minted}

\subsection*{Surrogate Model}

\texttt{mlr3mbo} supports any regression model from the mlr3 ecosystem, but as most acquisition functions depend on both the posterior mean and standard deviation, the learner must support \texttt{predict\_type = "se"} for posterior uncertainty estimation.
A complete list of supported learners can be found at \url{https://mlr-org.com/learners.html}.
If a learner does not natively support uncertainty estimation, bagging can be applied using the \texttt{mlr3pipelines}\cite{binder2021}  extension package to approximate the standard deviation.
Common choices for the surrogate learner include \texttt{regr.km}\cite{roustant2012}, a GP model with the usual kernels which is suitable for low- to medium-dimensional numeric search spaces, and \texttt{regr.ranger}\cite{wright2017}, a random forest model, which is appropriate for higher-dimensional, mixed, or hierarchical search spaces. 
Details on uncertainty estimation methods of the random forest are provided in the Supplement~\nameref{sec:uncertainty-estimation}.
For our example, we employ a GP with the Matérn~$5/2$ kernel.
The \texttt{SurrogateLearner} object is constructed with the \texttt{srlrn()} sugar function, which also takes the instance's archive as an argument.
When its \texttt{update()} method is called, it trains the SM with the observations stored in the archive.

\begin{minted}{r}
lrn_gp = lrn("regr.km", covtype = "matern5_2")
surrogate = srlrn(lrn_gp, archive = instance$archive)
surrogate$update()
\end{minted}

\subsection*{Acquisition Function}

Acquisition functions of class \texttt{AcqFunction} are stored in the \texttt{mlr\_acqfunctions} dictionary and can be constructed with the convenience function \texttt{acqf()}.
Table~\ref{tab:acqfunctions} lists the available acquisition functions.

We construct the EI acquisition function with the \texttt{acqf()} sugar function and then update it: 

\begin{minted}{r}
acq_function = acqf("ei", surrogate = surrogate)
acq_function$update()
acq_function$y_best # holds best y-value of the archive

# [1] -0.6792294
\end{minted}

Note that the \texttt{AcqFunction} object must also be updated in each iteration of the MBO loop, such that model- and data-dependent parts, e.g., the $f_{\mathrm{min}}$ value of the EI, or iteration-dependent scaling or decay control parameters can be taken into account.

\begin{figure}
    \centering
    \includegraphics[width=0.75\linewidth]{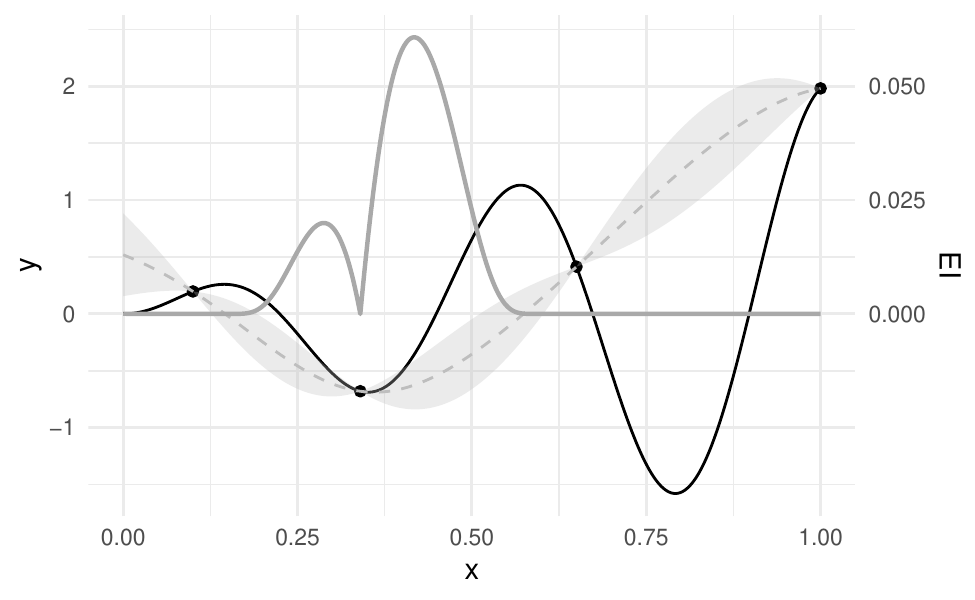}
    \caption{Expected Improvement (solid dark gray line) based on the mean (dashed gray line) and uncertainty (gray shaded area) prediction of the GP SM trained on an initial design of four points (black).}
    \label{fig:ei}
\end{figure}

We can now evaluate the acquisition function manually by passing a \texttt{data.table} of candidate points to its \texttt{\$eval\_dt()} method, as shown in Figure~\ref{fig:ei}. Note that we do this here for didactic purposes; in general it is not necessary to manually evaluate the acquisition function when using \texttt{mlr3mbo}.
We can see that the expected improvement is high in regions where the mean prediction (gray dashed lines) of the GP is low, balanced with high uncertainty (gray shaded area).

\subsection*{Acquisition Function Optimizer}

Acquisition function optimizers of class \texttt{AcqOptimizer} are stored in the \texttt{mlr\_acqopt} dictionary and can be constructed with the convenience function \texttt{acqo()}.
Table~\ref{tab:acqopt} lists the available acquisition function optimizers.
Here, we select the DIRECT\cite{jones1993} algorithm from the \texttt{nloptr} package.
In its default setting, the algorithm restarts $5 \cdot d$ times and runs at most for $100 \cdot d^2$ function evaluations, where $d$ is the number of dimensions of the search space.

\begin{minted}{r}
acq_optimizer = acqo("direct")

candidate = acq_optimizer$optimize()
candidate
#            x     acq_ei
#        <num>      <num>
# 1: 0.4172859 0.06074387
\end{minted}

\subsection*{Termination}

\label{sec:termination}

The termination of the optimization process is controlled by a user-specified criterion, as already shown in the first code chunk in this section.
It can be stopped after a certain number of function evaluations, a time limit, or a limit on the marginal improvement of the objective value over the previous iteration, for example.
\texttt{Terminator} objects can be constructed with the convenience function \texttt{trm()}.
Table~\ref{tab:terminators} lists the available terminators in the \texttt{bbotk} package.
Note that multiple criteria can also be combined with an ``and'' or ``or'' operation.

\subsection*{Predefined Loop Functions and Optimizers}
\label{sec:predefined-loop-functions}

Until now, we have described the core building blocks of \texttt{mlr3mbo} and how, by relying on these, a user can easily write their own flexible BO implementation in very few lines of code.
Section~\nameref{sec:loop_function} in the appendix collects all code blocks from the previous sections into a complete implementation of vanilla BO.
Alternatively, one can select a predefined \texttt{loop\_function} from the \texttt{mlr\_loop\_functions} dictionary.
These look very similar to the example in the appendix and run a predefined BO variant directly. 
Table~\ref{tab:loop-functions} summarizes the available loop functions.

\begin{table}[ht]
  \centering
  \caption{
    Loop functions are returned by \texttt{mlr\_loop\_functions\$get(key)}.
    Call \texttt{\$help()} to access the help page of the loop function e.g. \texttt{mlr\_loop\_functions\$get("bayesopt\_ego")\$help()}.}
  \label{tab:loop-functions}
  \begin{tabularx}{\textwidth}{lXl}
    \toprule
    Key             & Description                                          & Reference                        \\
    \midrule
    bayesopt\_ego    & Efficient global optimization                       & \protect\citeauthorsup{snoek2012}\protect\citeauthorsup{jones1998} \\
    bayesopt\_emo    & Multi-objective EGO                                 & \protect\citeauthorsup{huang2006}                \\
    bayesopt\_mpcl   & Multi-point constant liar                            & \protect\citeauthorsup{ginsbourger2008}\citeauthorsup{wang2020} \\
    bayesopt\_parego & Scalarization based multi-objective optimization    & \protect\citeauthorsup{knowles2006}               \\
    bayesopt\_smsego & Direct indicator based multi-objective optimization & \protect\citeauthorsup{beume2007}\citeauthorsup{ponweiser2008}  \\
    \bottomrule
  \end{tabularx}
\end{table}

Finally, \texttt{OptimizerMbo} can be used to assemble the building blocks described above into a single object for optimization, which is likely what most applied users want, and the MBO optimizer can be constructed via the sugar function \texttt{opt()}:

\begin{minted}{r}
bayesopt_ego = mlr_loop_functions$get("bayesopt_ego")
surrogate = srlrn(lrn("regr.km", covtype = "matern5_2",
  optim.method = "BFGS", control = list(trace = FALSE)))
acq_function = acqf("ei")
acq_optimizer = acqo("direct", maxeval = 10000L)

optimizer = opt("mbo",
  loop_function = bayesopt_ego,
  surrogate = surrogate,
  acq_function = acq_function,
  acq_optimizer = acq_optimizer)
\end{minted}

\texttt{OptimizerMbo} (or \texttt{opt("mbo")}) can also be called without any arguments to use the default configuration.
This configuration is adapted to the search space and codomain and is a good starting point for most tasks.
The default loop function for single-objective optimization is \texttt{bayesopt\_ego()} and for multi-objective optimization it is \texttt{bayesopt\_smsego()}.
For more details on the default configurations, see Section~\nameref{sec:mlr3mbo-configuration}.

Using these defaults, our example optimization can be implemented much more succinctly:

\begin{minted}{r}
instance = oi(objective, terminator = trm("evals", n_evals = 20))
design = data.table(x = c(0.1, 0.34, 0.65, 1))
instance$eval_batch(design)
optimizer$optimize(instance)

#            x  x_domain         y
#        <num>    <list>     <num>
# 1: 0.7921811 <list[1]> -1.577224
\end{minted}

\subsection*{Final Points}

\label{sec:final-points}

After termination, a concrete candidate must be returned from the archive. 
For noise-free problems, typically, the best observed point is recommended. 
For noisy problems, design points may have been evaluated a different number of times, so their objective values are known with varying certainty.
In such cases, the SM is usually employed to smooth out this noise. 
The user can control this behavior by passing a \texttt{ResultAssigner} as an argument to \texttt{OptimizerMbo}, which determines the final optimization result.
The default result assigner \texttt{ResultAssignerArchive} chooses the best point(s) across all evaluations in the archive.
An alternative result assigner is \texttt{ResultAssignerSurrogate}, which chooses the final point based on the prediction of an SM fit to all points in the archive.

\section{Advanced Features of \texttt{mlr3mbo}}

\label{sec:advanced-features}

\subsection*{Transformations}

\label{sec:transformations}

Transforming the features and target values of the SM can enhance its predictive performance, especially for a GP.
The \texttt{mlr3mbo} package provides a few pre-defined transformations; additional ones can be added easily by the user.
A simple transformation is to scale numeric features to the \([0, 1]\) range based on the search space boundaries.
The Kumaraswamy transformation on the features can be used to tackle non-stationarity~\cite{cowen-rivers2022}.
Furthermore, the target variable can be standardized to have zero mean and unit variance.
Logarithmic, Box-Cox, or Yeo-Johnson transformations can make the target more Gaussian and handle heteroscedasticity \cite{snelson2003, cowen-rivers2022}.
The transformation applied to the target values can be inverted before passing predictions to the acquisition function, or predictions may be retained on the transformed scale.
See Supplement \nameref{sec:output-transformations} for more information on output transformations.

The \texttt{SurrogateLearner} class supports transformations for both feature and target values; these are specified by passing \texttt{InputTrafo} or \texttt{OutputTrafo} objects to the \texttt{input\_trafo} and \texttt{output\_trafo} arguments.
\texttt{mlr3mbo} includes the \texttt{InputTrafoUnitcube} class, which scales the numeric features to the \([0, 1]\) range based on the search space boundaries, and the \texttt{OutputTrafoLog} class, which log-transforms target values after min-max scaling them to $[10^{-3}, 1]$.

The inverse transformation is automatically applied to predictions before they are passed to most acquisition functions. 
However, some acquisition functions, such as \texttt{AcqFunctionEILog} (see Section~\nameref{sec:acquisition-function}), operate directly on transformed targets.

In addition to transforming the feature and target values of the surrogate model, \texttt{paradox} supports transformations of points.
For example, the lower and upper bounds of a parameter can be specified on the log scale, and sampled values are exponentiated before being passed to the objective.

\subsection*{Random Interleaving}

\label{sec:random-interleaving}

Random interleaving samples a small number of proposed points uniformly at random rather than using the acquisition function \cite{bull2011,hutter2011}.
It is a simple yet effective technique that can improve the optimization process's performance by helping prevent the SM from getting stuck in local optima and strictly enforcing exploration, which can be particularly beneficial in the case of model mismatch\cite{ahmed2016}.
However, excessive random interleaving may reduce sample efficiency, so the frequency of random sampling should be carefully chosen.
Random interleaving is implemented in all pre-defined loop functions (see Section~\nameref{sec:predefined-loop-functions}).
The \texttt{random\_interleave\_iter} argument is passed to the loop function directly or via the \texttt{args} argument of \texttt{OptimizerMbo}, e.g.\ \texttt{opt("mbo", args = list(random\_interleave\_iter = 10))} for random sampling at every tenth iteration.

\subsection*{Warmstarting}

\label{sec:warmstarting}

Warmstarting incorporates known well-performing points into the initial design to guide the optimization process\cite{feurer2015}.
These points may originate from previous optimization runs or from domain knowledge, such as default parameter settings.
This provides the SM with prior information about promising regions of the search space, which can improve optimization efficiency.
In \texttt{mlr3mbo}, warmstarting is implemented by specifying a custom initial design.
Additionally, \texttt{mlr3mbo} supports online continuation: both \texttt{OptimizerMbo} and \texttt{TunerMbo} can be run with an existing archive containing configurations generated and evaluated by other optimizers.

\subsection*{Error Handling}

\label{sec:error-handling}

BO is more complex than many other optimization algorithms, and many different types of errors can disrupt it.
For example, fitting a GP can fail numerically, especially when two design points are close together.
To address such issues, \texttt{mlr3mbo} incorporates several built-in error handling and recovery mechanisms.
The \texttt{Surrogate} class provides a \texttt{catch\_errors} configuration parameter.
When set to \texttt{TRUE}, which is the default, this parameter ensures that all errors that occur during SM training or updates are caught and handled gracefully.
By default, the SM is encapsulated with a fallback learner.
If the SM fails, a random forest with 10 trees is used to predict the posterior mean and standard deviation.
The \texttt{AcqOptimizer} class similarly has a \texttt{catch\_errors} parameter to capture errors that occur during acquisition function optimization, whether due to surrogate prediction failures or errors in the acquisition optimizer itself.
If an error is detected at any of these steps, the loop function's default behavior is to fall back to sampling the next candidate point uniformly at random from the search space.

\subsection*{Mixed and Hierarchical Spaces in BO}

\label{sec:mixed-space-optimization}

Practical optimization problems frequently involve integer or categorical parameters, or even hierarchical structures, in which some parameters are conditionally active depending on the state of one or more parent parameters. 
Dealing with such spaces requires appropriate SMs and acquisition function optimizers.
Na\"ively treating integer-valued hyperparameters as real-valued ones and one-hot encoding categorical hyperparameters can lead to discrepancies between the acquisition function optimization and the optimization target, resulting in performance degradation \cite{garrido-merchan2020}.

Mixed and hierarchical spaces in GPs can be handled with appropriate kernels\cite{hutter2009,oh2019,kim2022,bergstra2011,levesque2017,horn2019}. 
\texttt{mlr3mbo} supports \texttt{PyTorch}\cite{ansel2024} GPs and kernels, which provide appropriate support for mixed spaces.
Alternatively, the GP can be replaced by models such as random forests\cite{hutter2009,hutter2011}, factorized multilayer perceptrons\cite{schilling2015}, or specialized density estimation algorithms\cite{bergstra2011}. 
These models are not implemented in \texttt{mlr3mbo}, but they could be added by implementing corresponding \texttt{Learner} classes in the \texttt{mlr3} ecosystem, for example via R or Python packages.
Random forests from the \texttt{ranger} package\cite{wright2017} are often considered to be the most effective choice for mixed-hierarchical search spaces\cite{hutter2011}.
Inactive parameters in hierarchical search spaces are encoded as missing values, which are handled natively by \texttt{ranger}.
Instead of choosing a suitable surrogate model for a specific search space, it is also possible to preprocess the inputs so that a model that does not natively support certain parameter types can still be used.
When using \texttt{mlr3mbo}, \texttt{mlr3pipelines}\cite{binder2021} can be used to impute missing values or to encode categorical hyperparameters numerically, enabling the use of standard GP models even for complicated search spaces.

For acquisition function optimization in mixed spaces, available approaches include stochastic LS\cite{hutter2011}, combinations of LS and gradient-based optimization\cite{garrido-merchan2020}, or probabilistic reparameterizations\cite{daulton2022}.
Other extensions use bandit algorithms\cite{ru2020} to handle categorical variables, or trust-region methods that scale BO to high-dimensional and mixed-integer spaces, such as TuRBO\cite{eriksson2019}, BAxUS\cite{papenmeier2022}, and Bounce\cite{papenmeier2023}.
For mixed spaces, \texttt{mlr3mbo} directly supports both RS and LS.
LS handles hierarchical dependencies by mutating only active parameters and checking the dependency conditions of the search space in topological order.
The Supplement~\nameref{sec:local-search} provides more details on our LS implementation.
While more sophisticated approaches are not supported by \texttt{mlr3mbo} out of the box, the package's modular design makes it possible to add custom BO loop functions that incorporate these concepts.

\subsection*{Multi-Objective Optimization}

\label{sec:multi-objective-optimization}

Multi-objective optimization addresses problems in which several objective functions, $f(\xv) = (f_1(\xv), \dots, f_k(\xv))$, are optimized simultaneously\cite{ehrgott2005}.
Unlike the single-objective case, there is no natural total order in $\mathbb{R}^k$ for $k \geq 2$.
Instead, solutions are compared using the concept of \emph{Pareto dominance}.
A point $\xv$ is said to Pareto-dominate another point $\tilde \xv$ (denoted $\xv \preceq \tilde \xv$) if $f_i(\xv) \leq f_i(\tilde \xv)$ for all $i = 1, \ldots, k$ and there exists at least one $j$ such that $f_j(\xv) < f_j(\tilde \xv)$.
A solution is \emph{non-dominated} if it is not dominated by any other point in the search space.
The set of all non-dominated points, $P = \{\tilde \xv \in \X \mid \nexists\, \xv \in \X : \xv \preceq \tilde \xv\}$, is called the \emph{Pareto set}, and its image under $f$ is the \emph{Pareto front}.
The primary goal in multi-objective optimization is to approximate the true Pareto set or the Pareto front as closely as possible.

The definition of the optimization problem is handled by the \texttt{bbotk} package and
multi-objective problems are defined with the \texttt{bbotk::OptimInstanceBatchMultiCrit} class.
The codomain of such an instance has multiple dimensions, and the objective function must return a vector of values for each input vector.
The \texttt{oi()} function automatically creates a multi-objective instance when the codomain includes more than one parameter.

\texttt{mlr3mbo} offers scalarization-based algorithms and direct indicator-based approaches for multi-objective optimization\cite{horn2015,karl2023}.
Scalarization-based algorithms such as ParEGO\cite{knowles2006} use projection and weighting techniques to reduce the multi-objective problem to a single-objective problem in each iteration, e.g.,
by Tchebycheff scalarization in ParEGO.
This approach is implemented in \texttt{mlr3mbo} as the \texttt{bayesopt\_parego()} loop function.
Direct indicator-based approaches like SMS-EGO\cite{beume2007} fit an SM for each objective and combine the predictions to a single scalar performance indicator.
SMS-EGO uses the hypervolume improvement as the acquisition function (\texttt{acqf("smsego")}) and is available in \texttt{mlr3mbo} as the \texttt{bayesopt\_smsego()} loop function.
A more complex direct indicator-based approach is the expected hypervolume improvement (EHVI) acquisition function\cite{emmerich2016}, implemented in \texttt{mlr3mbo} as \texttt{acqf("ehvi")}.
Further details and code examples for multi-objective optimization can be found in the \texttt{mlr3} book\cite{schneider2024}.

\section{Parallel Bayesian Optimization}
\label{sec:parallel}

The high computational cost of evaluating the black-box target function makes parallelization an important extension to accelerate the BO process.
\texttt{mlr3mbo} provides two general strategies for parallelization: multi-point batch proposal and asynchronous BO.

\subsection*{Multi-Point Batch Proposal}
\label{sec:multi-point-batch-proposal}

The BO process is inherently sequential, but evaluations can be parallelized by proposing multiple candidate points simultaneously to be evaluated by different worker processes.
Several approaches support this.
Multi-point expected improvement ($q$-EI)~\cite{ginsbourger2010} measures the expected improvement when evaluating a set of $q$ points simultaneously.
Another approach penalizes the acquisition function around already selected points~\cite{gonzalez2016}.

\texttt{mlr3mbo} implements constant liar~\cite{ginsbourger2010} via the \texttt{bayesopt\_mpcl()} loop function.
Here, the first point in $\xv^{(t+1, 1)}$ is obtained by ordinary EI maximization.
We then preliminarily impute a fake value $y^{(t + 1, 1)}$, update the SM, and generate $\xv^{(t + 1, 2)}$ again by EI maximization.
We proceed analogously for all further points in the batch.
Common choices for the imputed value are taken from the range between the minimum and the maximum of the y-values of the current archive $\mathcal{D}_t$ or the predicted posterior mean $\hat\mu(\xv^{(t+1, j)})$.
Our implementation is not restricted to EI but also works with other acquisition functions.
Parallel workers are started using the \texttt{future}~\cite{bengtsson2021} or \texttt{mirai}~\cite{gao2025} packages in R.
These packages can start workers locally (\texttt{future::multisession}, \texttt{future::multicore}, and \texttt{mirai::daemons}) or on remote machines (\texttt{future::cluster} and \texttt{mirai::daemons}).

\subsection*{Asynchronous Model-Based Optimization}
\label{sec:asynchronous-mbo}

Multi-point batch proposals are inefficient when the runtimes of points in a batch are highly heterogeneous, as the next batch can be proposed only after all evaluations have completed. 
To address this limitation, asynchronous approaches have been proposed.
One of the earliest works on asynchronous BO proposed sampling synthetic results from the SM\cite{snoek2012}.
This prevents the optimizer from selecting the same candidate point twice during acquisition function optimization.
Another approach introduced asynchronous BO via Thompson sampling~\cite{kandasamy2018} and draws random functions from the GP posterior and optimizes each sample separately to obtain a new candidate point.
Another notable approach is PLAyBOOK~\cite{alvi2019}, which applies penalties to the acquisition function around currently running points.

Asynchronous decentralized BO is a more recent approach~\cite{egele2023} for large-scale optimization on high-performance computing clusters.
Proposing multiple points in a centralized process can become a bottleneck, particularly when fitting the SM or optimizing the acquisition function is computationally expensive.
In asynchronous decentralized BO, each worker independently performs sequential BO and communicates its results asynchronously to other workers via a shared database.
Each worker fits its SM to the data available in the database and proposes new candidate points.
To ensure diversity among proposals, points currently being evaluated are imputed, akin to the constant liar approach, and the LCB acquisition function is used with different values of $\lambda$.

\texttt{mlr3mbo} currently supports the asynchronous decentralized BO architecture.
Asynchronous BO in \texttt{mlr3mbo} uses the same modular architecture as the synchronous variant and reuses many of the same building blocks.
The class \texttt{OptimizerAsyncMbo} with the \texttt{oi\_async} sugar construction function encapsulates the asynchronous optimization process.
Users can specify the SM, acquisition function, and acquisition function optimizer.
A key distinction in the asynchronous case is the use of stochastic acquisition functions~\cite{egele2023}, such as \texttt{AcqFunctionStochasticEI} or \texttt{AcqFunctionStochasticLCB}.
This facilitates proposing different candidate points by different workers.
The evaluation of the initial design also differs from the synchronous version -- it is created in the main process and distributed to the workers.
The \texttt{rush} package~\cite{becker2025} provides the shared database used for communication between workers.
The following example demonstrates how to set up and run asynchronous MBO in \texttt{mlr3mbo}.

\begin{minted}{r}
rush::rush_plan(n_workers = 2)

instance = oi_async(
  objective = objective,
  terminator = trm("evals", n_evals = 100))

acq_function = acqf("stochastic_cb",
  min_lambda = 1,
  max_lambda = 10)

optimizer = opt("async_mbo",
  design_size = 25,
  surrogate = surrogate,
  acq_function = acq_function,
  acq_optimizer = acq_optimizer)

optimizer$optimize(instance)

#            x  x_domain         y
#        <num>    <list>     <num>
# 1: 0.7918348 <list[1]> -1.577244
\end{minted}

\section{Hyperparameter Optimization}

\label{sec:hpo}

HPO aims to identify the optimal hyperparameter configuration for an ML system to optimize its predictive performance \cite{feurer2019,bischl2023,franceschi2025}.
In HPO, each candidate configuration is evaluated by training the learner and estimating its generalization performance, typically using a resampling technique such as cross-validation.
Evaluating a single configuration can be computationally expensive, especially for large datasets or models that are expensive to train.
The objective function that maps hyperparameter configurations to performance is generally unknown and lacks gradient information, making HPO a prototypical expensive black-box optimization problem.
This is a well-supported use case: \texttt{mlr3mbo} integrates seamlessly with the \texttt{mlr3tuning}\cite{becker2025a} package to provide efficient HPO via the \texttt{TunerMbo} class.
Analogously to \texttt{oi()} and \texttt{opt()}, which construct optimization instances and optimizers, \texttt{ti()} creates tuning instances and \texttt{tnr()} constructs tuners.
This tuner wraps \texttt{OptimizerMbo} and allows users to apply BO to hyperparameter search spaces.
The following example demonstrates the use of \texttt{mlr3mbo} for tuning the hyperparameters of a support vector machine (SVM) on the sonar dataset.
It is common for model fitting or prediction for the tuned learner to fail with an error.
We recommend using encapsulation and fallback learners to mitigate such problems, as described in the \texttt{mlr3} book\cite{bischl2024} in Chapter 10.
Both \texttt{cost} and \texttt{gamma} are tuned on the logarithmic scale, so that the values reported in the result are on the log scale and are transformed back to the original scale before being passed to the learner.

\begin{minted}{r}
library(mlr3tuning)

tuner = tnr("mbo")

lrn_svm = lrn("classif.svm", 
  kernel = "radial",
  type = "C-classification",
  cost  = to_tune(1e-5, 1e5, logscale = TRUE),
  gamma = to_tune(1e-5, 1e5, logscale = TRUE)
)

instance = ti(
  task = tsk("sonar"),
  learner = lrn_svm,
  resampling = rsmp("cv", folds = 3),
  measure = msr("classif.ce"),
  terminator = trm("evals", n_evals = 25)
)

tuner$optimize(instance)

#        cost     gamma learner_param_vals  x_domain classif.ce
#       <num>     <num>             <list>    <list>      <num>
# 1: 11.19589 -3.553767          <list[4]> <list[2]>  0.1588682
\end{minted}

\subsection*{AutoML and CASH}

\label{sec:automl}
Combined algorithm selection and hyperparameter optimization (CASH) treats the selection of the learning algorithm and the optimization of its hyperparameters as a single optimization problem.
CASH underpins most AutoML frameworks\cite{thornton2013,feurer2022} that consider entire ML pipelines, which include data preprocessing, ML algorithm selection, and hyperparameter tuning.
The corresponding search spaces are often large and contain complex mixed-space structures, including hierarchical dependencies.

\begin{figure}[ht]
  \centering
  \includegraphics[width=0.75\linewidth]{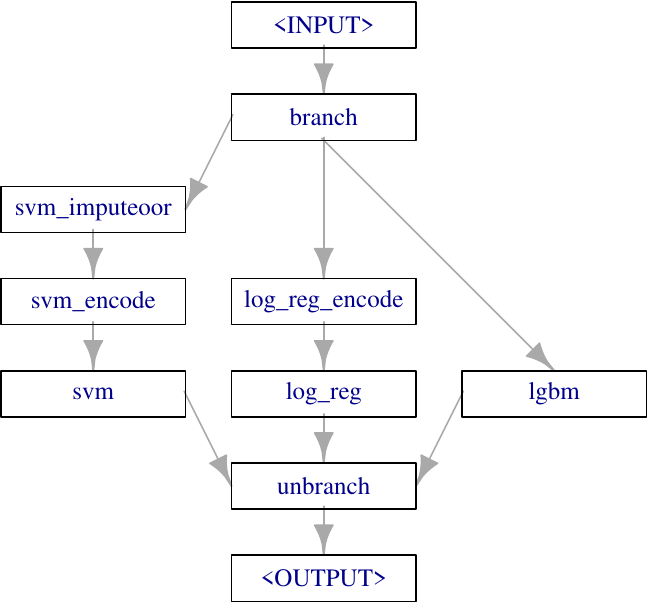}
  \caption{Graph of our example AutoML system.}
  \label{fig:graph}
\end{figure}

We demonstrate the construction of a simple AutoML system using the \texttt{mlr3pipelines} \cite{binder2021} package.
The system consists of three learning algorithms: LightGBM, SVM, and logistic regression.
These algorithms have different capabilities for handling missing data and categorical features.
Consequently, we add algorithm-specific preprocessing steps to encode categorical features and impute missing data if needed.
The preprocessing steps and learners are combined into a graph structure that branches into the different algorithms.
Figure~\ref{fig:graph} illustrates the graph structure of our example AutoML system.

\begin{minted}{r}
  library(mlr3verse)
  
  lgbm = lrn("classif.lightgbm", id = "lgbm")
  svm =  po("imputeoor", id = "svm_imputeoor") %>>% 
    po("encode", method = "one-hot", id = "svm_encode")  %>>% 
    lrn("classif.svm", type = "C-classification", kernel = "radial", id = "svm")
  log_reg = po("imputeoor", id = "log_reg_imputeoor") %>>%
    po("encode", method = "one-hot", id = "log_reg_encode")  %>>%
    lrn("classif.log_reg", id = "log_reg")

  graph_learner = as_learner(ppl("branch",
    list(lgbm = lgbm, svm = svm, log_reg = log_reg)), id = "automl",
    prediction_type = "prob")
 \end{minted}

The \texttt{as\_learner()} function converts the graph into a learner that can be tuned like any other learner.
The graph learner has a joint hyperparameter search space with all algorithm- and preprocessing-specific hyperparameters and a categorical hyperparameter \texttt{branch.selection} that determines which branch of the graph is selected.
We define the search space for the graph learner, including the dependency on the branch.

\begin{minted}{r}
  search_space = ps(
    lgbm.learning_rate = p_dbl(lower = 0.01, upper = 0.3,
      depends = branch.selection == "lgbm"),
    lgbm.max_depth = p_int(lower = 3, upper = 10,
      depends = branch.selection == "lgbm"),
    svm.cost = p_dbl(lower = 0.1, upper = 10,
      depends = branch.selection == "svm"),
    svm.gamma = p_dbl(lower = 0.001, upper = 1,
      depends = branch.selection == "svm"),
    branch.selection = p_fct(levels = c("lgbm", "svm", "log_reg"))
  )
\end{minted}

We employ the BO tuner to optimize the example AutoML system with this complex search space.
When the search space contains dependencies, the tuner automatically selects a random forest as the SM.
  
\begin{minted}{r}
instance = ti(
  task = tsk("pima"),
  learner = graph_learner,
  resampling = rsmp("cv", folds = 3),
  measure = msr("classif.auc"),
  terminator = trm("evals", n_evals = 100),
  search_space = search_space
)

tuner = tnr("mbo")
tuner$optimize(instance)

# lgbm.learning_rate lgbm.max_depth  svm.cost  svm.gamma branch.selection    auc
# 1:              NA             NA      0.97       0.03              svm   0.85
\end{minted}

For brevity, we omit the code for nested resampling, which is essential for obtaining unbiased generalization performance estimates of the AutoML system \cite{bischl2023}.
The performance estimate from the code above is based on the inner resampling and is over-optimistic.
Nested resampling can be implemented with \texttt{AutoTuner} and the \texttt{resample()} function.
More information on nested resampling and pipelines can be found in the \texttt{mlr3} book\cite{bischl2024}.

\section{Related Work}
\label{sec:related-work}

Several BO software frameworks have been proposed in recent years, each employing and supporting different SMs and acquisition strategies to effectively optimize expensive black-box functions.
The predecessor of the \texttt{mlr3mbo} package is \texttt{mlrMBO}\cite{bischl2018}. 
The current package was redesigned to be more modular, to follow object-oriented design principles more closely, and to integrate with the \texttt{mlr3} ecosystem\cite{lang2019}.
The \texttt{mlrMBO} package has been widely used across diverse application areas; for example in actuarial science\cite{chen2023}, agriculture\cite{roscher-ehrig2024}, environmental science\cite{hough2020}, hydrology\cite{yang2021}, chemistry\cite{rodriguez-perez2020,yong2025}, geoinformatics\cite{merghadi2018}, and medicine\cite{smirniotis2024,ianevski2019,browaeys2020}.

Several other BO packages are available for R.
\texttt{rBayesianOptimization}\cite{yan2016} provides a basic implementation with GPs as SMs.
\texttt{ParBayesianOptimization}\cite{wilson2018} supports parallel evaluations, but is also limited to GP SMs. 
\texttt{GPareto}\cite{binois2019} supports multi-objective optimization, again limited to GP SMs.
The \texttt{tune}~\cite{kuhn2025} package from the \texttt{tidymodels} ecosystem provides HPO functionality and supports a GP SM.
Users can choose between EI, PI, and LCB as acquisition functions, which are optimized using a space-filling design.

In addition to R, several popular BO implementations exist in Python.
\texttt{SMAC3} \cite{lindauer2022} is widely used for hyperparameter tuning and general black-box optimization.
It provides similar features to \texttt{mlr3mbo} but puts more emphasis on out-of-the-box usability, particularly for users without prior experience with BO.
\texttt{HEBO} \cite{cowen-rivers2022} is a popular alternative in Python and the winning submission for the NeurIPS 2020 Black-Box Optimization Challenge.
\texttt{Ax} is another alternative that emphasizes general optimization rather than HPO. 
It directly supports GPU acceleration through \texttt{BoTorch} but also goes beyond BO by allowing users to use Bandit algorithms to optimize for example purely categorical search spaces.
Other popular Python packages for BO include \texttt{Hyperopt}\cite{bergstra2011}, \texttt{Optuna}\cite{akiba2019}, and \texttt{Dragonfly}\cite{kandasamy2020}.
Many of these packages provide functionality and flexibility comparable to \texttt{mlr3mbo}.

More recently, several Python-based BO frameworks have emerged that target domain-specific or applied optimization settings.
\texttt{NUBO}\cite{diessner2025} is a lightweight framework based on the \texttt{Torch} ecosystem, emphasizing transparency and accessibility for non-expert users optimizing physical experiments, supporting sequential and parallel evaluations over mixed search spaces.
\texttt{BoFire}\cite{durholt2024} and \texttt{BayBE}\cite{fitzner2025} both build on \texttt{BoTorch}~\cite{balandat2020} and are designed for chemistry and materials science applications, providing built-in chemical encodings, multi-objective optimization, and extensive constraint handling; \texttt{BayBE} additionally supports transfer learning across related optimization campaigns.
\texttt{BOA}\cite{scyphers2024} wraps \texttt{Ax} and \texttt{BoTorch} behind a text-based (YAML or JSON) configuration interface, enabling language-agnostic usage and built-in parallelization for HPC environments.
While these tools primarily target applied users who want effective BO without needing to understand its internals, \texttt{mlr3mbo} additionally targets BO researchers and advanced practitioners by exposing the full algorithmic pipeline in a modular and transparent way.
It is agnostic to the choice of surrogate model class and tightly integrates with a broader machine learning ecosystem for hyperparameter optimization, enabling users to freely combine arbitrary learners, acquisition functions, and optimization strategies.

Implementations of BO are available in many other languages, for example \texttt{BayesOpt}\cite{martinez-cantin2014} for C++, \texttt{BayesianOptimization} for Julia, and \texttt{bayesopt} in Matlab.
Various black-box optimization methods have been studied with respect to their tunability and the impact of their configuration parameters on optimization performance.
Lindauer et al.\ (2019)\cite{lindauer2019} meta-optimized BO using SMAC.
Moosbauer et al.\ (2022)\cite{moosbauer2022} configured both the optimization algorithm itself and its configuration parameters using an integrated approach. 

\section{Benchmarks}
\label{sec:benchmarks}

\subsection*{\texttt{mlr3mbo} Configuration}
\label{sec:mlr3mbo-configuration}

Given the high configurability of \texttt{mlr3mbo}, our goal is to identify a default configuration that performs well and robustly across a wide range of scenarios.

\paragraph{Experimental Setup}
We constructed a comprehensive meta-search space with many possible configurations supported by the package.
As BO has different requirements for purely numeric and mixed-hierarchical problems, and because an optimal default configuration could differ between these regimes, we defined corresponding search spaces (Tables~\ref{tab:mlr3mbo-search-space-numeric} and \ref{tab:mlr3mbo-search-space-mixed}).
The numeric parameters of \texttt{mlr3mbo} were discretized into a small set of representative values to obtain purely categorical search spaces.
To obtain the optimal configuration, we used a greedy, coordinate-wise local search that iteratively evaluates all configurations in a one-parameter exchange neighborhood of the current incumbent and accepts the best configuration found at each iteration.
This procedure changes one parameter coordinate at a time and is equivalent to a coordinate descent (CD) algorithm with greedy coordinate selection.
This has the advantage that the meta-optimization process becomes more interpretable:
For each configuration chosen during the optimization procedure, we can determine which changes have the largest impact on performance, and we can quantify the magnitude of improvements in terms of average performance across the benchmark set. Moreover, we can start from an extremely simple or standard configuration in which more complex or non-standard settings are only added or changed if empirically justified.

Each configuration was evaluated on the YAHPO-SO~v1 benchmark suite from the YAHPO Gym package\cite{pfisterer2022}.
YAHPO Gym provides surrogate-based benchmarks of HPO tasks with varying metrics, dimensionalities, and complexities, and the v1 suite comprises a diverse subset of the most faithful SMs from the full collection.
We followed the authors' recommendation to repeat the evaluation of each instance 30 times with different random seeds.
We use a budget of $\lceil 100 + 40 \sqrt{d} \rceil$ for $d$ dimensions.
The numeric and mixed-hierarchical instances used in our experiments are listed in Tables~\ref{tab:numeric_instances} and~\ref{tab:mixed_instances}, together with the respective dataset names and search space dimensions.

Although all instances use accuracy as the performance metric, the range of attainable values can differ substantially between instances.
We therefore define a performance measure that scores each configuration relative to RS, which, in our opinion, reflects practical relevance. 
We refer to this measure as the ``random search normalized score'' (RSNS).
A score of 0 corresponds to the performance of RS with $\lceil 100 + 40 \sqrt{d} \rceil$ evaluations, whereas a score of 1 corresponds to the performance of RS with 1{,}000{,}000 evaluations.
This construction follows the idea of using extensive RS as a common reference baseline\cite{turner2021}.
To obtain the reference value for RSNS of 1, we first ran RS with a budget of 1{,}000{,}000 evaluations on each YAHPO instance.
The performance of RS with a smaller budget, corresponding to RSNS 0, was then simulated by subsampling 30 of the smaller searches from the large RS trace. 
For each configuration, we compute the RSNS based on the mean performance across 30 independent runs with different random seeds to reduce noise.
Finally, we average the RSNS over all instances.
The coordinate descent procedure is as follows:

\begin{enumerate}
    \item \textbf{Initialization:} Set and evaluate the start configuration (see Table~\ref{tab:mlr3mbo-results-numeric} and \ref{tab:mlr3mbo-results-mixed}) and designate it as the incumbent (current best).
    \item \textbf{Candidate Generation:} For each parameter (coordinate), consider all possible values while keeping all other parameters fixed at their incumbent values.
    \item \textbf{Evaluation:}
    \begin{enumerate}
        \item Evaluate the performance of each candidate configuration on the instances, averaged over 30 runs, and calculate the RSNS.
        \item Average the RSNS over all instances.
    \end{enumerate}
    \item \textbf{Selection:} Identify the candidate configuration that has the best performance. If this configuration improves over the incumbent, use it as the new incumbent.
    \item \textbf{Iteration:} Repeat steps 2--4. The process terminates when the best candidate no longer improves over the incumbent.
\end{enumerate}

Dependent parameters are always evaluated jointly with all admissible values of their parent parameter.
For example, when the LCB acquisition function is selected, we evaluate it with all values of the trade-off parameter $\lambda$ from our pre-specified grid.

To quantify the variability of the RSNS, we estimate its standard deviation for the final configuration via a Monte Carlo sampling procedure.
For this configuration, we generated 3{,}000 independent evaluations on each instance.
We then repeatedly drew 30 evaluations without replacement from these 3{,}000 runs for each instance, computed the RSNS, and averaged it as during our CD evaluation.
Repeating this procedure 1{,}000 times yielded an empirical distribution of the RSNS estimator, for which we compute the standard deviation.
While this procedure was applied only to the best configuration, it is still informative about the typical scale of RSNS variability across configurations.

See Section~\nameref{sec:technical-details} for more information regarding further, more technical settings.
All code to reproduce our experiments is available on zenodo at \url{https://doi.org/10.5281/zenodo.18223637}. 
All runs were made fully reproducible by fixing random seeds.
The experiments were conducted on the National Center for Atmospheric Research's Derecho cluster\cite{derecho2023}, which provides 128 AMD Milan cores and 256 GB shared memory per node and runs SUSE Enterprise Linux 15.4.0.
Running the coordinate descent on the numeric and mixed-hierarchical search spaces required approximately 7.7 CPU-years of computation.

\paragraph{Results}

\begin{table}[ht]
  \centering
  \caption{Start and final configuration of the CD on numeric instances.
    For all evaluated configurations, see Table~\ref{tab:archive_numeric}.
    Parameters that are only relevant for the random forest surrogate model are omitted.}
  \label{tab:mlr3mbo-results-numeric}
  \begin{tabular}{lll}
    \toprule
    Parameter              & Start value & Final value \\
    \midrule
    Input trafo            & None          & None        \\
    Output trafo           & None          & Log         \\
    Init                   & Random        & Random      \\
    Init size              & 0.25          & 0.05        \\
    Random interleave      & 0             & 0           \\
    Surrogate              & GP            & GP          \\
    Kernel                 & Gauss         & Matérn~$3/2$  \\
    Nugget                 & 0             & $10^{-8}$        \\
    Scaling                & FALSE         & FALSE       \\
    Acquisition function   & EI            & LCB         \\
    $\lambda$              & (not active)              & 3           \\
    Optimizer & RS 1000       & CMA-ES      \\
    $\epsilon$ decay       & FALSE         &  (not active)           \\
    $\lambda$ decay        &   (not active)            & FALSE       \\
    \bottomrule
  \end{tabular}
\end{table}

\begin{table}[ht]
  \centering
  \caption{Optimization path of the CD on numeric instances.}
  \label{tab:optimization-path-numeric}
  \begin{tabular}{rllr}
    \toprule
    Iteration & Parameter            & Value                                             & RSNS \\
    \midrule
              & Start configuration  & --                                                & 0.76 \\
    1         & Optimizer            & CMA-ES                                            & 0.99 \\
    2         & Output trafo         & Log                                               & 1.08 \\
    3         & Surrogate            & GP - Matérn~$3/2$ - Nugget $=0$ - No Scaling         & 1.14 \\
    4         & Acquisition function & LCB - $\lambda=3$                                 & 1.18 \\
    5         & Surrogate            & GP - Matérn~$3/2$ - Nugget $=10^{-8}$ - No Scaling & 1.18 \\
    6         & Init size            & 0.05                                              & 1.19 \\
    \bottomrule
  \end{tabular}
\end{table}

For purely numeric tasks, we started with a simple configuration that uses a GP SM with the EI acquisition function (Table~\ref{tab:mlr3mbo-results-numeric}), no input or output transformations, a large initial design, and a low-cost RS for acquisition function optimization. 
The start configuration achieved a mean RSNS of 0.76.
In the first iteration, the acquisition optimizer was changed to CMA-ES, which increased the mean RSNS to 0.99.
At this point, \texttt{mlr3mbo} (with its much smaller budget) performed, on average, on par with an RS with 1{,}000{,}000 evaluations.
In subsequent iterations, the output transformation was changed to a logarithmic transformation, the kernel was changed to a Matérn~$3/2$ kernel and a nugget of $10^{-8}$, and the acquisition function was changed to LCB with a $\lambda$ value of 3, yielding a mean RSNS of 1.18.
In later iterations, the RSNS varied around 1.18 without further systematic improvement (see Table~\ref{tab:archive_numeric}).
Decreasing the initial design size to 5\% of the budget performs comparably to 10\% and 25\% while yielding better anytime performance than competing frameworks, as shown in Section~\nameref{sec:performance-comparison}.
This is our default configuration for numeric tasks, which achieves a mean RSNS of 1.19.
The standard deviation of the RSNS of the final configuration is 0.01.

Replacing CMA-ES with RS or L-BFGS-B decreases the RSNS by at least 0.2, as shown in Table~\ref{tab:ablation-numeric}.
Only LS achieves an RSNS comparable to CMA-ES.
Changing the acquisition function from LCB decreases the RSNS in all cases.
Furthermore, removing the logarithmic output transformation reduces the RSNS by 0.21.
The type and size of the initial design have no substantial effect on the RSNS.

\begin{table}[ht]
  \centering
  \caption{Start and final configuration of the CD on mixed-hierarchical instances.
    For all evaluated configurations, see Table~\ref{tab:archive_mixed}.}
  \label{tab:mlr3mbo-results-mixed}
  \begin{tabular}{lll}
    \toprule
    Parameter            & Start value & Final value \\
    \midrule
    Input trafo          & None          & None        \\
    Output trafo         & None          & Log         \\
    Init                 & Random        & Random      \\
    Init size            & 0.25          & 0.05        \\
    Random interleave    & 0             & 0           \\
    Variance estimator   & ESD           & LTV         \\
    Trees                & 500           & 500         \\
    Acquisition function & EI            & LCB         \\
    $\lambda$            & (not active)              & 1           \\
    Optimizer               & RS 1000       & LS          \\
    $\epsilon$ decay     & FALSE         & (not active)            \\
    $\lambda$ decay      & (not active)              & FALSE       \\
    \bottomrule
  \end{tabular}
\end{table}

\begin{table}[ht]
  \centering
  \caption{Optimization path of the CD on mixed-hierarchical instances.}
  \label{tab:optimization-path-mixed}
  \begin{tabular}{rllr}
    \toprule
    Iteration & Parameter            & Value             & RSNS \\
    \midrule
              & Start configuration  & --                & 0.34 \\
    1         & Output trafo         & Log               & 0.50 \\
    2         & Optimizer            & LS                & 0.69 \\
    3         & Variance estimator   & LTV               & 0.78 \\
    4         & Init size            & 0.1               & 0.78 \\
    5         & Acquisition function & LCB - $\lambda=3$ & 0.80 \\
    6         & Acquisition function & LCB - $\lambda=1$ & 0.80 \\
    7         & Init size            & 0.05              & 0.82 \\
    \bottomrule
  \end{tabular}
\end{table}

For mixed-hierarchical tasks, we started the CD procedure with a random forest SM with the ESD variance estimator, again with EI, no input or output transformations, a large initial design, and a low-cost RS for acquisition function optimization, as shown in Table~\ref{tab:mlr3mbo-results-mixed}.
In the first iteration, a logarithmic output transformation was added, increasing the RSNS from 0.34 to 0.50.
In subsequent iterations, the acquisition optimizer was changed to LS, and the variance estimator of the SM was changed to LTV, yielding an RSNS of 0.78.
Further iterations slightly improved the RSNS to 0.82 by changing the acquisition function to LCB with $\lambda$ = 1 and reducing the initial design size to 5\% of the budget.
The last iteration did not improve the RSNS further (see Table~\ref{tab:archive_mixed}).
The standard deviation of the RSNS of the final configuration is 0.02.
Overall, the final default configuration achieves substantially higher relative performance on numeric instances (mean RSNS 1.19) than on mixed-hierarchical instances (mean RSNS 0.82), indicating that mixed-hierarchical tasks remain challenging for \texttt{mlr3mbo}.

Replacing LS with RS decreases the RSNS by 0.20, as shown in Table~\ref{tab:ablation-mixed}.
Another important parameter is the logarithmic transformation of the output.
Compared to the log-transformed setting, the RSNS is lower by 0.22 with no output transformation and by 0.08 with standardization.
Using only 10 trees instead of 500 trees in the random forest decreases the RSNS by 0.15.

\subsection*{Comparison to State-of-the-Art HPO Tools}
\label{sec:performance-comparison}

\paragraph{Experimental Setup}

We compared the new default configurations of \texttt{mlr3mbo} against \texttt{Ax} (v1.1.2)\cite{olson2025}, \texttt{HEBO} (v0.3.6)\cite{cowen-rivers2022}, \texttt{SMAC3} (v2.3.1)\cite{lindauer2022}, and \texttt{Optuna} (v4.5.0)\cite{akiba2019} on the YAHPO-SO v1 benchmark suite.
The instance set was the same as in Section~\nameref{sec:mlr3mbo-configuration}.
The evaluation budget was the YAHPO recommended budget of $\lceil 20 + 40 \sqrt{d} \rceil$ evaluations.
On the numeric instances, we compared \texttt{Ax}, \texttt{HEBO}, and \texttt{Optuna} in their default settings, \texttt{SMAC3} configured with the \texttt{SMAC4BB} and \texttt{SMAC4HPO} facades, and \texttt{mlr3mbo} in its new default configuration for numeric search spaces.
On the mixed-hierarchical instances, we compared \texttt{Optuna} in its default settings, \texttt{SMAC3} with the \texttt{SMAC4HPO} facade (as its \texttt{SMAC4BB} facade does not support complex search spaces), and \texttt{mlr3mbo} in its new default configuration for mixed spaces.
We restricted the acquisition function optimization budget to a maximum of 10{,}000 evaluations in \texttt{mlr3mbo} because the default of $100 \cdot d^2$ resulted in excessively long runtimes for high-dimensional instances.
\texttt{Ax} and \texttt{HEBO} do not support mixed-hierarchical search spaces for BO out of the box and were therefore excluded from this part of the comparison.
Each experiment was independently repeated 30 times with different random seeds.
We report both the performance over iterations and the final performance for each framework.
As before, we use RSNS as performance metric.
To compare the final performance of the frameworks across all instances, we use a Friedman test with post-hoc analysis as 
recommended by Demšar\cite{demsar2006}.

\paragraph{Results}

\begin{figure}[h]
  \centering
  \begin{subfigure}{0.45\textwidth}
    \includegraphics[width=\textwidth]{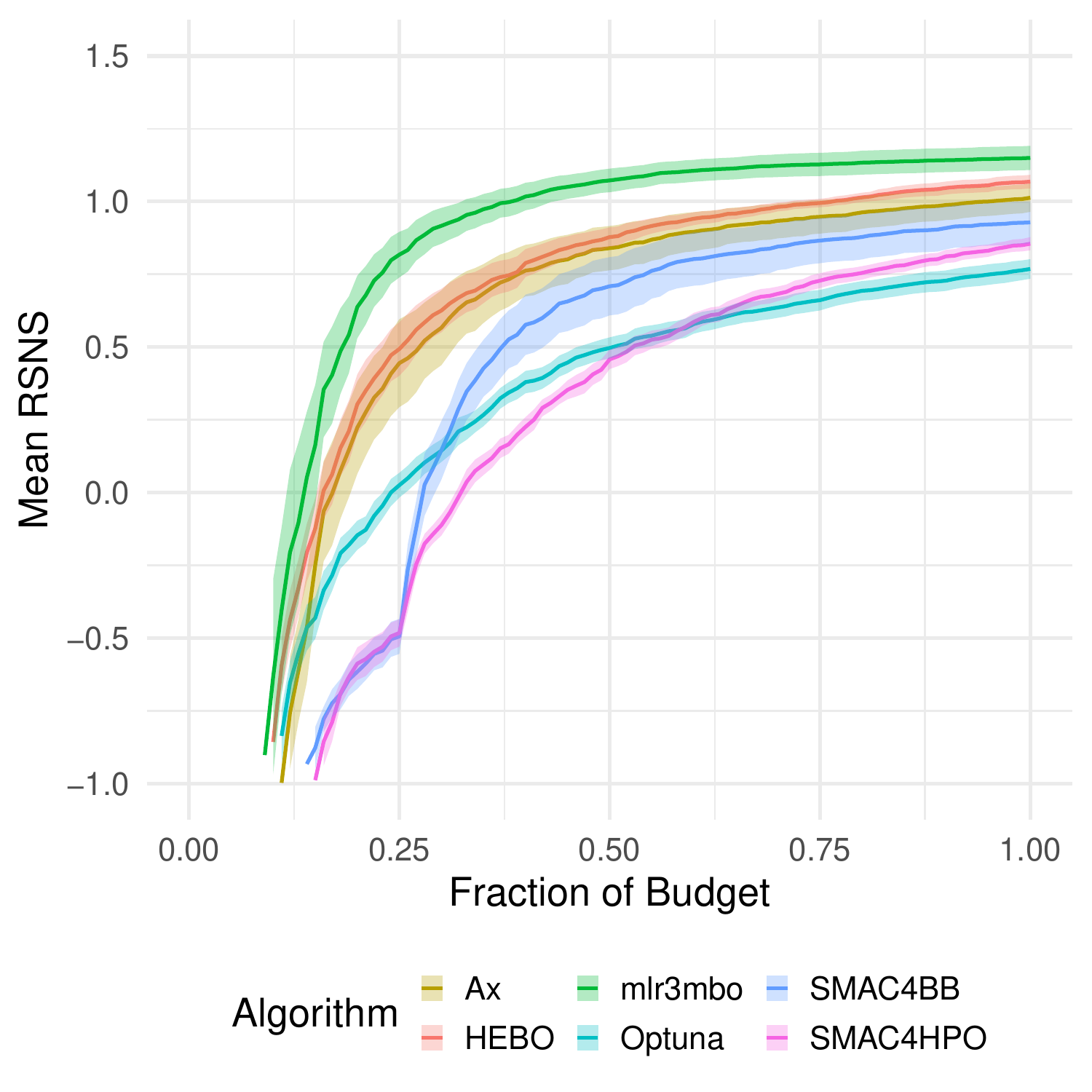}
    \caption{Numeric instances.}
    \label{fig:performance-pure-numeric-mean-meta-score}
  \end{subfigure}
  \begin{subfigure}{0.45\textwidth}
    \includegraphics[width=\textwidth]{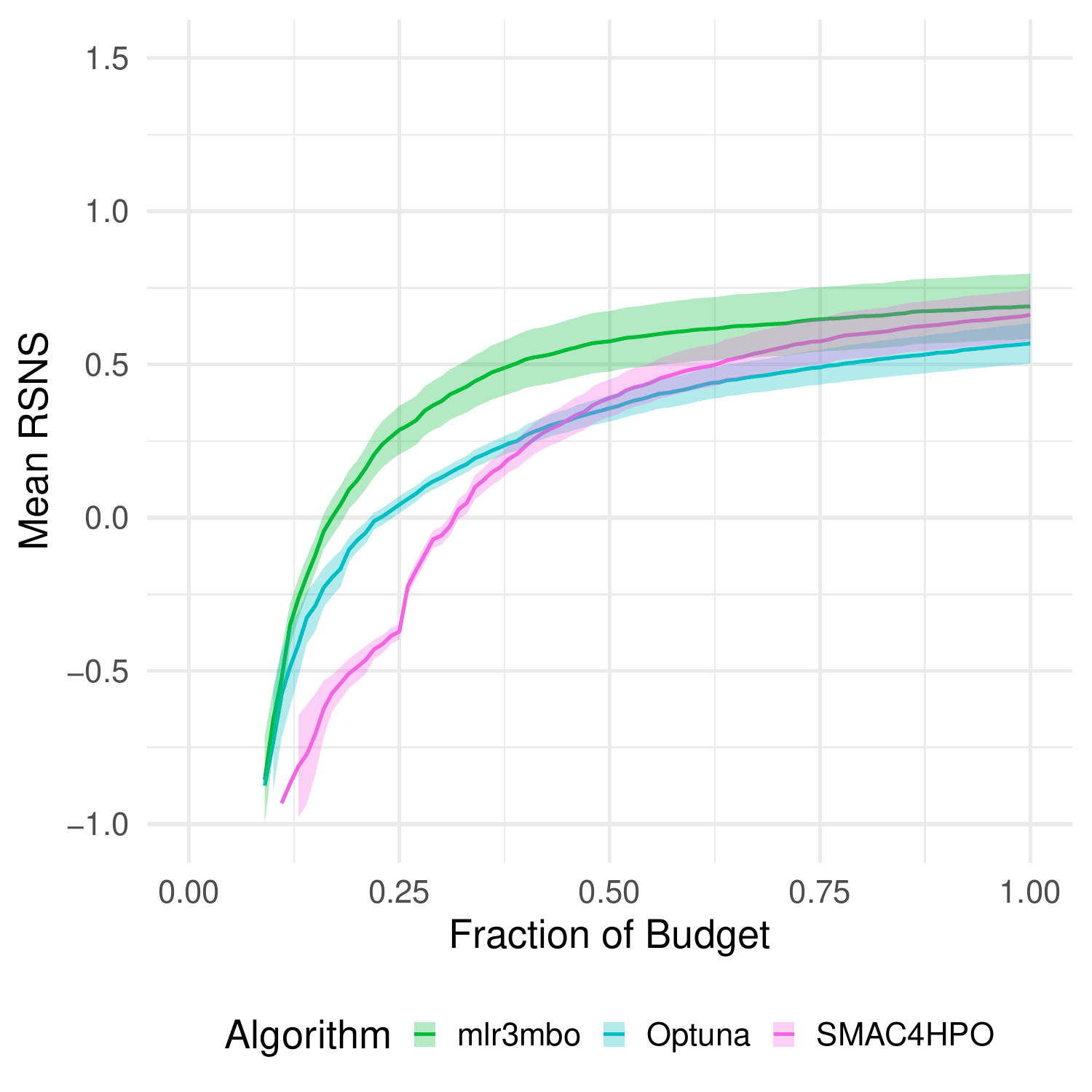}
    \caption{Mixed-hierarchical instances.}
    \label{fig:performance-mixed-deps-mean-meta-score}
  \end{subfigure}
  \caption{Optimization performance of \texttt{Ax}, \texttt{HEBO}, \texttt{SMAC4BB}, \texttt{SMAC4HPO}, \texttt{Optuna}, and \texttt{mlr3mbo} over iterations.
  Performance is measured as the mean RSNS over all instances.
  Shaded ribbons show the standard error of the mean RSNS across the instances.
  The curves start at different budget fractions because the initial design size is different for each framework.}
\end{figure}

\begin{figure}[ht]
  \centering
  \begin{subfigure}{0.45\textwidth}
    \includegraphics[width=\textwidth]{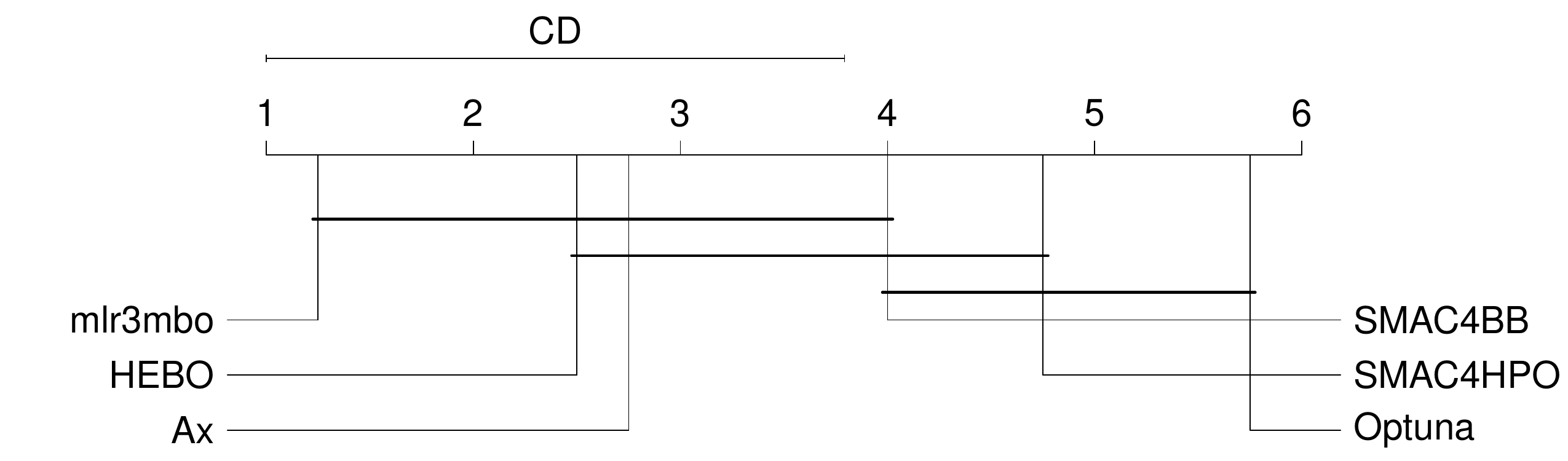}
    \caption{Numeric instances.}
    \label{fig:performance-pure-numeric-critical-difference}
  \end{subfigure}
  \begin{subfigure}{0.45\textwidth}
    \includegraphics[width=\textwidth]{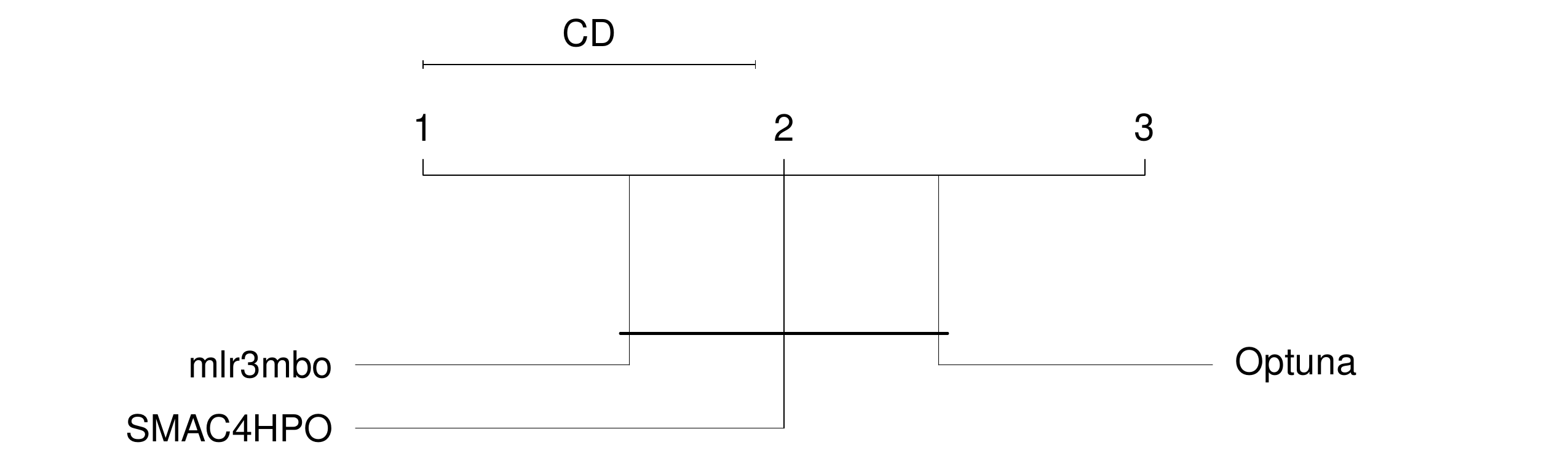}
    \caption{Mixed-hierarchical instances.}
    \label{fig:performance-mixed-deps-critical-difference}
  \end{subfigure}
  \caption{Critical difference diagrams of the average ranks of the final performance across all instances.
  The Friedman test is significant at $\alpha = 0.05$ for numeric instances.
  The horizontal bars indicate that frameworks connected by a bar do not differ significantly according to the Nemenyi post-hoc test at $\alpha = 0.05$.
  The Friedman test is close to significance at $\alpha = 0.05$ for mixed-hierarchical instances.
  The critical difference plot is consistent with this result, as all frameworks are connected by a horizontal bar.}
\end{figure}

The performance on numeric instances is shown in Figure~\ref{fig:performance-pure-numeric-mean-meta-score}.
\texttt{mlr3mbo} achieves the highest mean RSNS among \texttt{Ax}, \texttt{HEBO}, and both \texttt{SMAC} variants from the start, although the rank-based comparison in Figure~\ref{fig:performance-pure-numeric-critical-difference} indicates parity rather than significant superiority over the strongest baselines.
Both \texttt{SMAC3} facades use a larger initial design budget than the other frameworks and start model-based optimization later.
\texttt{SMAC4BB} quickly catches up after 50\% of the evaluation budget has been spent.
We expected \texttt{SMAC4HPO} to perform worse on the numeric instances because its random forest SM is not tailored to purely numeric search spaces.
\texttt{Ax}, \texttt{HEBO}, and \texttt{mlr3mbo} exhibit strong anytime performance on almost all instances because they already perform well with a small initial design (Figure S1).
Figure~\ref{fig:performance-pure-numeric-critical-difference} shows the critical difference diagram of the average ranks of the final performance across all instances.
Lower ranks correspond to better performance, and frameworks connected by a horizontal bar do not differ significantly according to the Nemenyi post-hoc test at $\alpha = 0.05$.
The diagram indicates that \texttt{mlr3mbo} performs at a comparable level to \texttt{Ax}, \texttt{HEBO}, and \texttt{SMAC4BB} (no significant differences), while \texttt{SMAC4HPO} and \texttt{Optuna} achieve a significantly higher average rank than \texttt{mlr3mbo}.

Figure~\ref{fig:performance-mixed-deps-mean-meta-score} shows the performance comparison on mixed-hierarchical spaces.
Again, \texttt{mlr3mbo} has better anytime performance than \texttt{SMAC4HPO}.
\texttt{Optuna} starts stronger than \texttt{SMAC4HPO} but falls behind after 50\% of the evaluation budget has been spent.
The lower mean performance of \texttt{Optuna} is mainly driven by its weaker scores on the YAHPO instances corresponding to HPO problems involving a single ML model (non-CASH-type problems) (Figure S2).
All frameworks perform similarly on the high-dimensional NASBench (neural architecture search) instance (Figure S2), and on the AutoML pipeline instances, the best framework changes multiple times.
Figure~\ref{fig:performance-mixed-deps-critical-difference} shows the critical difference diagram for mixed-hierarchical instances; no two frameworks differ significantly.

\begin{figure}
  \centering
  \begin{subfigure}{0.45\textwidth}
    \includegraphics[width=\textwidth]{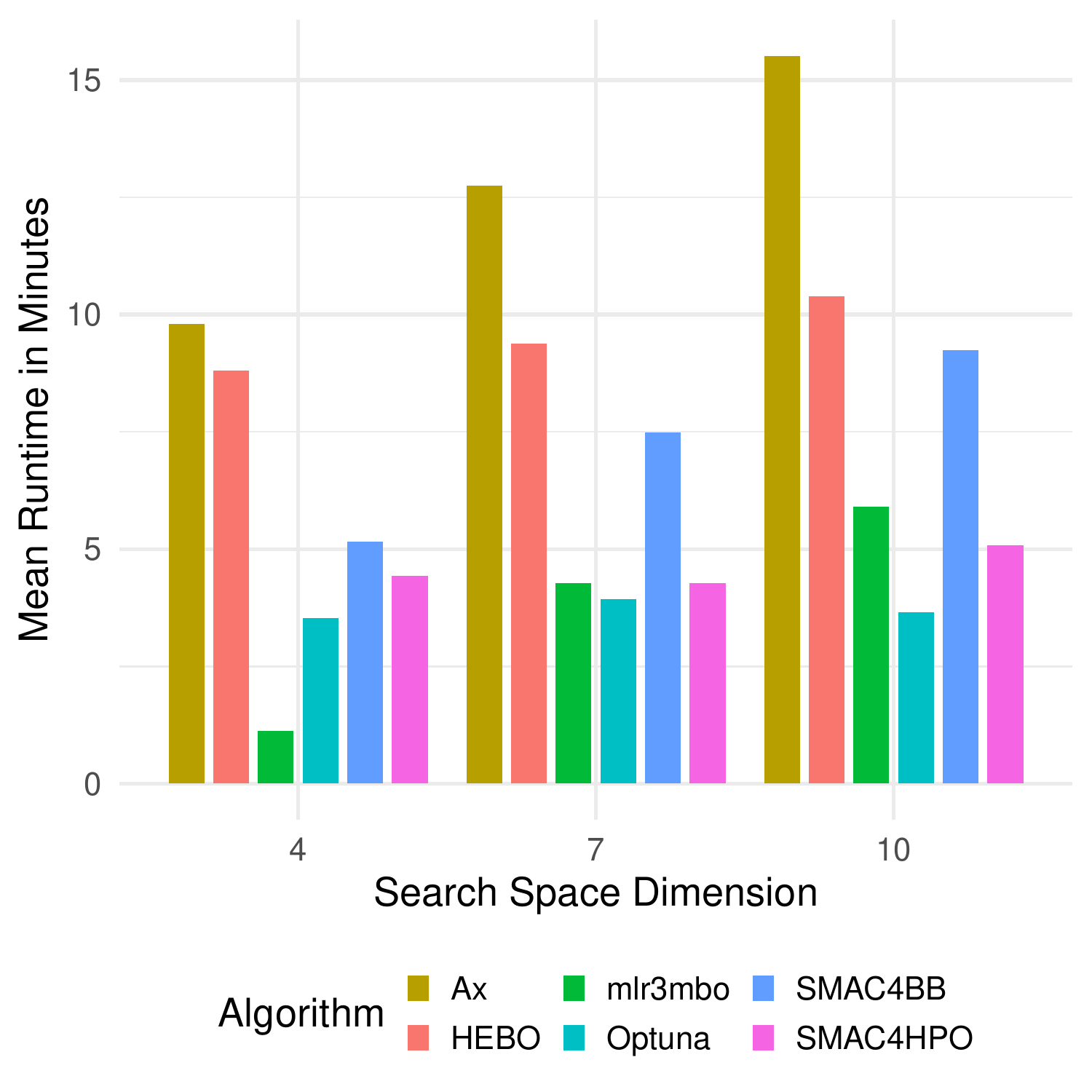}
    \caption{Numeric instances.}
    \label{fig:runtime-pure-numeric}
  \end{subfigure}
  \begin{subfigure}{0.45\textwidth}
    \includegraphics[width=\textwidth]{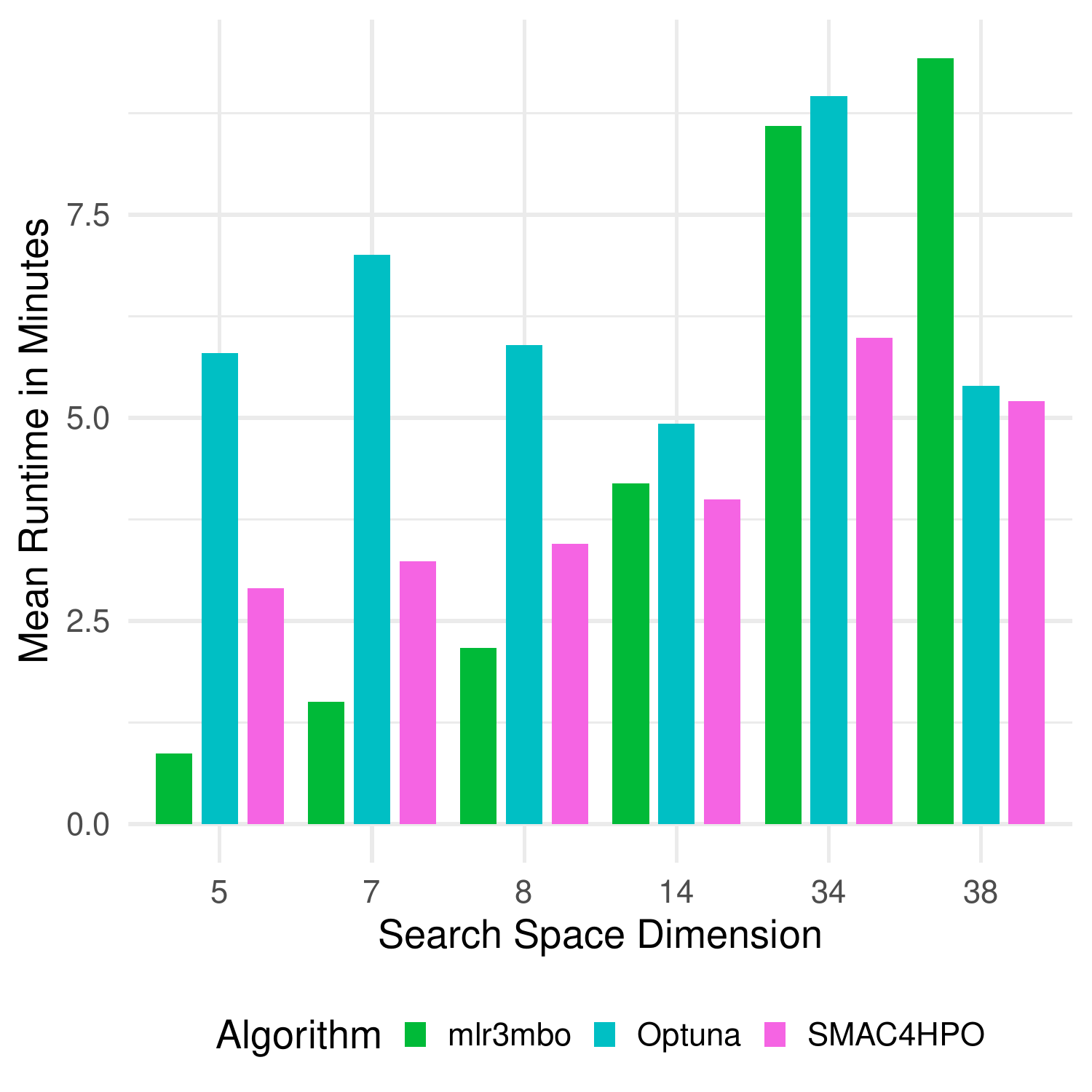}
    \caption{Mixed-hierarchical instances.}
    \label{fig:runtime-mixed-deps}
  \end{subfigure}
  \caption{Runtime in minutes of \texttt{Ax}, \texttt{HEBO}, \texttt{SMAC4BB}, \texttt{SMAC4HPO}, \texttt{Optuna}, and \texttt{mlr3mbo}.}
\end{figure}

As the runtime overhead of BO toolkits (mainly stemming from surrogate modeling and acquisition function optimization) can be substantial and is a further relevant criterion for practical use, we also measured it.
We emphasize that the quantity reported here is deliberately the toolkit overhead, not the end-to-end wall-clock time of a BO run on a real-world objective.
Since YAHPO instances are evaluated through SMs themselves, evaluating an instance at a point is nearly instantaneous, and the measured runtime of a BO run on YAHPO is almost exclusively toolkit overhead.
We view this as a feature rather than a limitation of the setup: end-to-end runtime on any real-world objective can be made arbitrarily large or small by choosing a more or less expensive target function, and that cost is identical for every framework on the same problem.
What differs between frameworks is the overhead that the user ``pays'' on top of every expensive evaluation in exchange for the framework's optimization behavior, i.e.\ how efficiently the toolkit realizes its chosen BO variant.
The comparison below should therefore be read in this sense: it quantifies framework efficiency at fixed BO-performance level, not the total time of a hypothetical application run.
On numeric instances, the runtime of \texttt{mlr3mbo} is consistently lower than that of the competing frameworks (Figure~\ref{fig:runtime-pure-numeric}).
For low-dimensional mixed-hierarchical search spaces, \texttt{mlr3mbo} is faster than the other frameworks (Figure~\ref{fig:runtime-mixed-deps}).
On 38-dimensional instances, however, \texttt{mlr3mbo} requires approximately double the runtime of the competitors.
The higher runtime of \texttt{mlr3mbo} in these settings is attributable to the larger budget devoted to acquisition function optimization in high-dimensional spaces.
We would expect this increased effort to translate into better performance, but this is not the case (see Figure S2).

\subsection*{Discussion and Conclusions}
\label{sec:discussion}

Our experiments show that, with carefully chosen defaults, \texttt{mlr3mbo} matches the performance of established state-of-the-art BO frameworks on numeric and mixed-hierarchical YAHPO Gym benchmarks, 
while offering good runtime behavior.
Our results underscore the importance of strong acquisition function optimizers with sufficiently large budgets and a surrogate tailored to the search space's structure.
CMA-ES and LS are the best choices for optimizing acquisition functions on numeric and mixed-hierarchical instances, respectively.
Using a GP on numeric instances is the best choice, whereas on mixed-hierarchical instances, a random forest with a variance estimator based on LTV works well.
Our default configurations are consistent with empirical studies demonstrating that GPs perform best on numeric hyperparameter spaces \cite{eggensperger2015}.
Our finding that CMA-ES and LS perform well as acquisition function optimizers, and that log-transformed targets help, is also consistent with prior studies \cite{bergstra2011, wang2013, hutter2011, lindauer2019}.

The new default configuration of \texttt{mlr3mbo} for numeric instances matches the configurations of \texttt{Ax} and \texttt{SMAC4BB} in that it uses a GP SM.
Differences arise in the acquisition functions, where \texttt{SMAC3} and \texttt{Ax} rely on EI, while \texttt{mlr3mbo} uses LCB.
We also observed that \texttt{mlr3mbo} performs much worse when using L-BFGS-B as the acquisition optimizer, even though \texttt{Ax} successfully relies on L-BFGS-B.
This may be because \texttt{Ax} uses automatic differentiation to compute gradients of the acquisition function, whereas \texttt{mlr3mbo} estimates gradients via finite differences.
L-BFGS-B was a strong candidate in the first CD iteration, indicating that the parameters chosen in later iterations interact poorly with L-BFGS-B and that our final configuration is better tuned to CMA-ES (and LS) than to gradient-based optimization.
Interestingly, \texttt{mlr3mbo} benefits from a simple random initial design, whereas \texttt{Ax} and \texttt{SMAC4BB} rely on space-filling designs.

Both \texttt{SMAC4HPO} and \texttt{mlr3mbo} use a random forest SM and LS as the acquisition function optimizer for mixed-hierarchical search spaces.
However, \texttt{SMAC3} uses only 10 trees in its random forest and estimates the variance using ESD, whereas \texttt{mlr3mbo} uses 500 trees and LTV.
The original \texttt{SMAC} algorithm switched from ESD to LTV at a later stage, but \texttt{SMAC3} reverted to ESD in its current implementation~\cite{hutter2014, lindauer2022}.
When configured with only 10 trees, \texttt{mlr3mbo} performs notably worse (Table~\ref{tab:ablation-mixed}), indicating that very small forests provide insufficiently stable uncertainty estimates for effective acquisition optimization.
While \texttt{SMAC3} also supports LCB, its default HPO configuration uses LogEI.

\paragraph{Limitations and Future Work}
While CD offers a simple and interpretable approach to configuring \texttt{mlr3mbo}, it also has important limitations.
CD optimizes one parameter at a time while keeping the others fixed.
This can lead to suboptimal configurations when parameter interactions are strong.
For example, improving the SM to obtain a better uncertainty estimate only helps if the acquisition function exploits that uncertainty.
In such a procedure, the final configuration is also influenced by the starting configuration -- which one may or may not like, depending on whether one wants to analyze differences with respect to a reasonable baseline start point.
Moreover, the RSNS metric is noisy. 
The CD procedure may therefore move in random directions in the last iterations, although we
average over a large number of instances and 30 repetitions. 
We mainly handled this by letting CD run ``long enough'' and fixing a stopping point by ``eyeballing'' -- one could have been more statistically formal about this (but properly handling sequential tests is not trivial). 

The CD search and the performance comparison were conducted on the same set of benchmark instances.
This is consistent with how most BO and HPO frameworks report their performance: HEBO\cite{cowen-rivers2022}, SMAC3\cite{lindauer2022}, BoTorch/Ax\cite{balandat2020}, and Optuna\cite{akiba2019} are likewise developed and benchmarked on the same families of problems (e.g.\ Bayesmark, HPOBench, or standard synthetic suites) rather than on strictly held-out external families.
Nevertheless, the setup could in principle lead to overfitting of the \texttt{mlr3mbo} configuration to the YAHPO instances.
The benchmark suite is, however, large and diverse, covering AutoML pipelines and a variety of ML algorithms across different dimensionalities, which we believe substantially reduces the risk of overfitting to suite-specific properties; the CD study additionally makes the empirical basis for each configuration choice explicit, rather than leaving the defaults as undocumented maintainer-chosen values.
A fully external evaluation on a held-out benchmark family would further strengthen the generality of these claims and is a useful direction for future work.

Using surrogate-based benchmarks for HPO problems enables extensive experimentation at low computational cost.
However, surrogates can only approximate the behavior of the underlying ML algorithms.
A major difference from real HPO problems is that real HPO problems are noisy, depending on resampling splits and random seeds, whereas surrogate benchmarks return deterministic results. 
Further work on real, non-surrogate HPO problems is needed to evaluate the performance of the frameworks in fully realistic settings.

Furthermore, while \texttt{mlr3mbo} and the competing frameworks support various options for batch-parallel or asynchronous BO, the present study focused exclusively on sequential BO.
Future research should extend this.

Another natural direction for future work is the support of multi-fidelity and multiple-information-source BO.
The classical idea of leveraging cheaper, lower-fidelity approximations of the objective alongside the expensive high-fidelity evaluation dates back to Kennedy and O'Hagan\cite{kennedy2000} and has since been formalized in the multi-information-source framework of Poloczek et al.\cite{poloczek2017}.
In hyperparameter optimization, multi-fidelity approaches such as BOHB\cite{falkner2018} have become a de facto standard, and there is growing recent interest in their use for cost- and energy-aware AutoML\cite{candelieri2024}.
Native support of these methods would require extending the surrogate model and acquisition function interfaces of \texttt{mlr3mbo} to handle fidelity/source variables and their associated cost models, and is high on our development roadmap.

\newpage

\section*{RESOURCE AVAILABILITY}

\subsection*{Lead contact}

Requests for further information and resources should be directed to and will be fulfilled by the lead contact, Bernd Bischl (bernd.bischl@stat.uni-muenchen.de).

\subsection*{Materials availability}

This study did not generate new materials.

\subsection*{Data and code availability}

\begin{itemize}
  \item All original code and results have been deposited at Zenodo\cite{becker2026a} under \url{https://doi.org/10.5281/zenodo.18223637} and are publicly available as of the date of publication.
  This Zenodo record links to a GitHub release of the accompanying repository at \url{https://github.com/mlr-org/mbo_config} for the experimental analysis.
  \item The \texttt{mlr3mbo} package is available at \url{https://github.com/mlr-org/mlr3mbo/} and on CRAN.
  \item Any additional information required to reanalyze the data reported in this paper is available from the lead contact upon request.
\end{itemize}

\section*{ACKNOWLEDGMENTS}

We would like to acknowledge high-performance computing support from the Derecho system (doi:10.5065/qx9a-pg09) provided by the NSF National Center for Atmospheric Research (NCAR), sponsored by the National Science Foundation. In addition, we thank Matthias Feurer and Valentin Margraf for helpful discussions about the project and feedback on the paper draft.

\section*{AUTHOR CONTRIBUTIONS}

Conceptualization, B.B., M.BE. and L.S.; 
code, L.S., M.BE., B.B.; 
empirical analysis, M.BE., L.S., B.B.; 
writing--original draft, M.BE., B.B.;
writing--review \& editing, B.B., M.BE., M.BI., L.K.;
funding acquisition, B.B.; 
computational resources, L.K.;
supervision, B.B., L.K.

\section*{DECLARATION OF INTERESTS}

The authors declare no competing interests.

\section*{DECLARATION OF GENERATIVE AI AND AI-ASSISTED TECHNOLOGIES}

The authors used GPT-5.1 and Grammarly to proofread for clarity, coherence, grammar, and punctuation. 
GPT-5.1 was also sometimes used to search for potentially missing references (Google-style searches and general literature search).
Generative AI tools were used to assist with some initial code for experiments and in the package; nearly all code was written manually.
The authors reviewed and edited all content and take full responsibility for it.

\newpage

\section*{SUPPLEMENTAL INFORMATION INDEX}

\begin{description}
  \item Table S1. Acquisition functions
  \item Table S2. Acquisition function optimizers 
  \item Table S3. Terminators
  \item Table S4. Search space for purely numeric instances
  \item Table S5. Search space for mixed-hierarchical instances
  \item Table S6. YAHPO Gym numeric instances
  \item Table S7. YAHPO Gym mixed-hierarchical instances
  \item Table S8. Ablation study on numeric configuration
  \item Table S9. Ablation study on mixed-hierarchical instances
  \item Table S10. Coordinate descent archive on numeric instances
  \item Table S11. Coordinate descent archive on mixed-hierarchical instances
  \item Table S12. Software packages referenced in this paper with their latest release versions
  \item Text S1. Uncertainty estimation methods
  \item Text S2. Loop function
  \item Text S3. Output Transformations
  \item Text S4. Local Search
  \item Text S5. Technical Details
  \item Figure S1. Mean RSNS per numeric instance.
  \item Figure S2. Mean RSNS per mixed-hierarchical instance.
\end{description}

\newpage

\input{supplements/acquisition_functions}
\input{supplements/optimizers}

\input{supplements/terminators}
\input{supplements/numeric_search_space}
\input{supplements/mixed_search_space}
\input{supplements/numeric_instances}
\input{supplements/mixed_instances}
\input{supplements/numeric_ablation}
\input{supplements/mixed_ablation}
\input{supplements/software-versions}

\newpage

\input{supplements/uncertainty_estimation}

\newpage

\input{supplements/loop_function}

\newpage

\input{supplements/output_transformations}

\newpage

\input{supplements/local_search}

\newpage

\input{supplements/technical_details}

\newpage

\includepdf[landscape=true,pages=-]{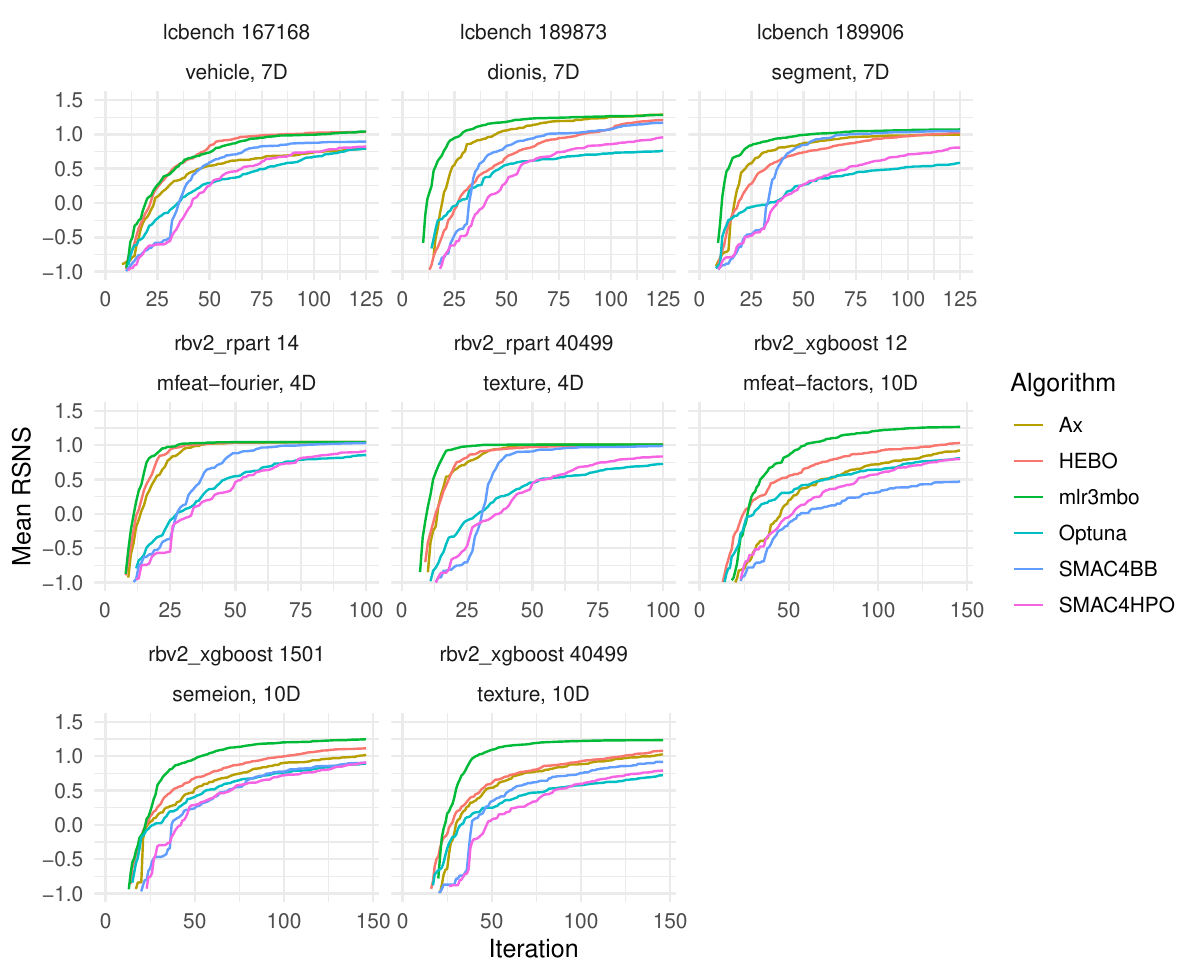}

\newpage

\includepdf[landscape=true,pages=-]{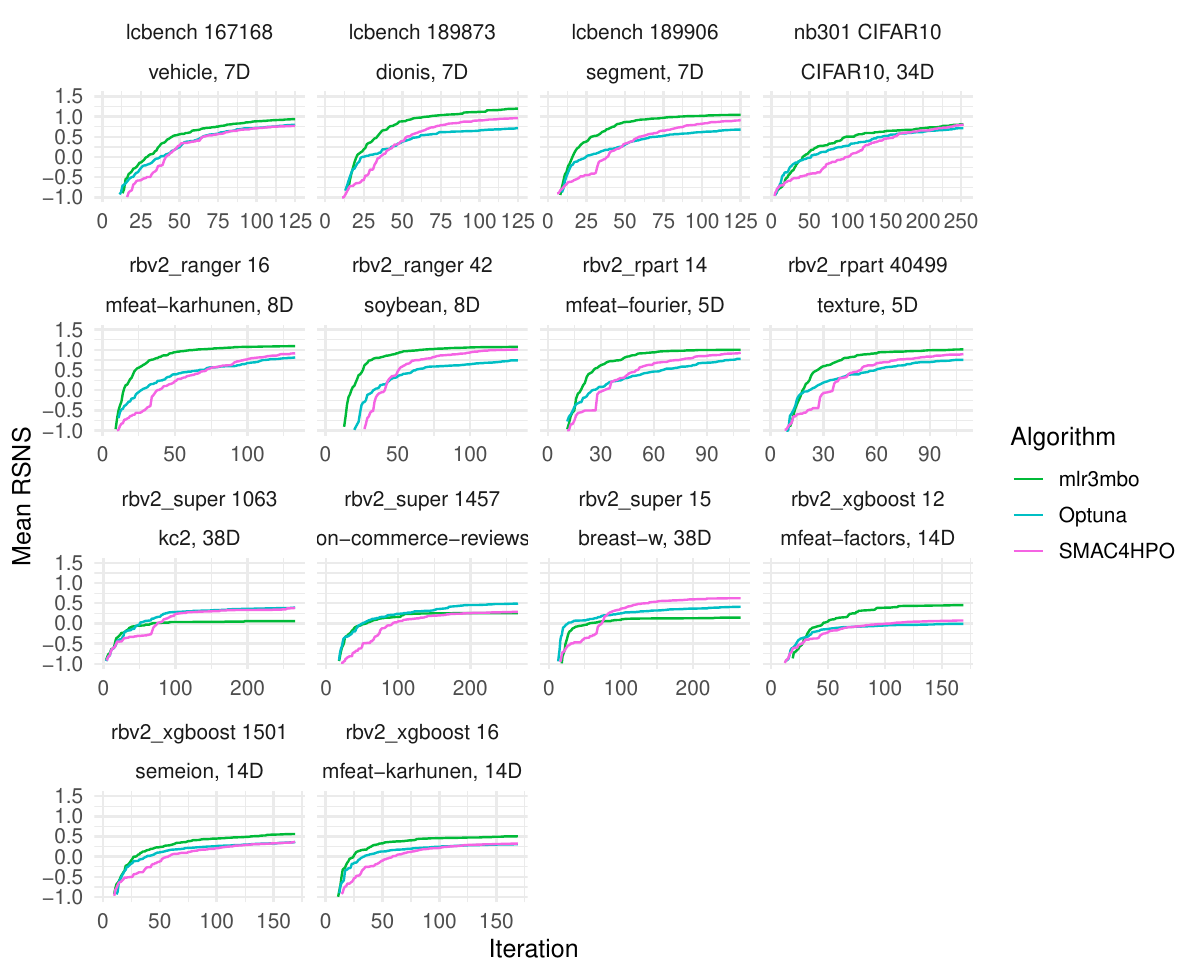}

\newpage

See archive\_numeric.xlsx for the complete archive of numeric instances.

\begin{table}[h]
  \caption{Archive of numeric instances.}
  \label{tab:archive_numeric}
\end{table}

\newpage

See archive\_mixed.xlsx for the complete archive of mixed-hierarchical instances.

\begin{table}[h]
  \caption{Archive of mixed-hierarchical instances.}
  \label{tab:archive_mixed}
\end{table}

\newpage

\newpage

\bibliography{references}

\bigskip

\newpage

\end{document}

%% file: helperFunctions.tex
\usepackage{amsmath, amsfonts}
\usepackage{mathtools}
\usepackage{xspace}

\newcommand{\R}{\ensuremath{\mathbb{R}}}                   
\DeclareMathOperator*{\argmin}{arg\,min}                   

\newcommand{\X}{\ensuremath{\mathcal{X}}\xspace}           
\newcommand{\xv}{\ensuremath{\boldsymbol{x}}\xspace}       
\newcommand{\fh}{\ensuremath{\hat{f}}\xspace}              
\newcommand{\fx}{\ensuremath{f(\xv)}\xspace}               

\usepackage{pifont}


%% file: supplements/acquisition_functions.tex
\begin{table}[H]
  \centering
  \caption{Acquisition Functions.
    Acquisition functions are constructed with the \texttt{acqf(key)}.
    Call \texttt{\$help()} to access the help page of the acquisition function e.g. \texttt{acqf("ei")\$help()}.
    The help page describes the formula of the acquisition function and the parameters that can be set.}
  \label{tab:acqfunctions}
  \begin{tabularx}{\textwidth}{l>{\RaggedRight\arraybackslash}Xl}
    \toprule
    Key            & Description                                                                                                                                 & Reference                   \\
    \midrule
    \texttt{aei}            & Augmented expected improvement for noisy optimization. $c$ controls the exploration-exploitation trade-off                                  & \protect\citeauthorsup{huang2006}    \\
    \texttt{cb}             & Lower / upper confidence bound. $\lambda$ controls the exploration-exploitation trade-off                                                   & \protect\citeauthorsup{snoek2012}    \\
    \texttt{ehvi}           & Expected hypervolume improvement for multi-criteria optimization                                                                            & \protect\citeauthorsup{emmerich2016} \\
    \texttt{ehvigh}         & Expected hypervolume improvement via GH quadrature for multi-criteria optimization                                                          & \protect\citeauthorsup{rahat2022}    \\
    \texttt{ei}             & Expected improvement                                             & \protect\citeauthorsup{jones1998}    \\
    \texttt{ei\_log}        & Expected improvement. Assumes that the surrogate models on log scale                                                            &                          \protect\citeauthorsup{hutter2011}\citeauthorsup{watanabe2024}   \\
    \texttt{eips}           & Expected improvement per second. Balances performances and computation time                                                                 & \protect\citeauthorsup{snoek2012}    \\
    \texttt{mean}           & Posterior mean. Mean of the surrogate model                                                                                                 &                             \\
    \texttt{multi}          & Multiple acquisition functions. Combines multiple acquisition functions                                                                     &                             \\
    \texttt{pi}             & Probability of improvement.                                                                                                                 & \protect\citeauthorsup{kushner1964}  \\
    \texttt{sd}             & Posterior standard deviation. Standard deviation of the surrogate model                                                                     &                             \\
    \texttt{smsego}         & S-metric selection evolutionary global optimization for multi-criteria optimization. Maximizing the expected improvement in the hypervolume & \protect\citeauthorsup{beume2007}    \\
    \texttt{stochastic\_cb} & Stochastic lower / upper confidence bound for asynchronous optimization. Samples $\lambda$ from a distribution                              & \protect\citeauthorsup{snoek2012}    \\
    \texttt{stochastic\_ei} & Stochastic expected improvement for asynchronous optimization. Samples $\epsilon$ from a distribution                                       & \protect\citeauthorsup{jones1998}    \\
    \bottomrule
  \end{tabularx}
\end{table}

%% file: supplements/optimizers.tex
\begin{table}[H]
  \centering
  \caption{Acquisition function optimizers.
    Optimizers are constructed with the \texttt{acqo(key)}.
    Call \texttt{\$help()} to access the help page of the acquisition function optimizer e.g. \texttt{acqo("direct")\$help()}.
    The help page describes the parameters that can be set.}
  \label{tab:acqopt}
  \begin{tabularx}{\textwidth}{llXl}
    \toprule
    Key            & Search space type      & Default settings                                                        & Reference           \\
    \midrule
    \texttt{direct}         & Numeric                & $5 \cdot d$ restarts and $100 \cdot d^2$ function evaluations           & \citeauthorsup{jones1993}    \\
    \texttt{lbfgs}          & Numeric                & $5 \cdot d$ restarts and $100 \cdot d^2$ function evaluations           & \citeauthorsup{byrd1995}     \\
    \texttt{cmaes}          & Numeric                & ABIPOP with unlimited restarts and $100 \cdot d^2$ function evaluations & \citeauthorsup{hansen2001}   \\
    \texttt{random\_search} & Mixed and dependencies & $100 \cdot d^2$ iterations                                              &  \\
    \texttt{local\_search}  & Mixed and dependencies & 10 neighbors and $100 \cdot d^2$ iterations                             & \citeauthorsup{hutter2011}   \\
    \bottomrule
  \end{tabularx}
\end{table}

%% file: supplements/terminators.tex
\begin{table}[H]
  \centering
  \caption{\texttt{Terminator} classes in the \texttt{bbotk} package. 
  Terminators are constructed with the \texttt{trm(key)} function.
  Call \texttt{\$help()} to access the help page of the terminator e.g. \texttt{trm("stagnation")\$help()}.
  The help page describes the parameters that can be set.} 
  \label{tab:terminators}
  \begin{tabular}{ll}
    \toprule
    Key                     & Description                                                    \\
    \midrule
    \texttt{clock\_time}             & Clock time exceeds a given limit                              \\
    \texttt{combo}                   & A combination of terminators is met                           \\
    \texttt{evals}                   & Number of evaluations exceeds a given limit                   \\
    \texttt{none}                    & Optimization is finished                                      \\
    \texttt{perf\_reached}           & Performance reaches a given level                             \\
    \texttt{run\_time}               & Run time exceeds a given limit                                \\
    \texttt{stagnation}              & Performance has not improved for a given number of iterations \\
    \texttt{stagnation\_batch}       & Performance has not improved for a given number of batches    \\
    \texttt{stagnation\_hypervolume} & Hypervolume stagnates                                         \\
    \bottomrule
  \end{tabular}
\end{table}

%% file: supplements/numeric_search_space.tex
\begin{table}[H]
  \centering
  \caption{Search space for purely numeric instances.}
  \label{tab:mlr3mbo-search-space-numeric}
  \begin{tabular}{lllll}
    \toprule
    Parameter            & Values                          & Dependency     & Default \\
    \midrule
    Input trafo          & None, unitcube                  & --             & None    \\
    Output trafo         & None, standardize, log          & --             & None    \\
    Init                 & Random, LHS, Sobol              & --             & Random  \\
    Init size            & 0.05, 0.10, 0.25                & --             & 0.25    \\
    Random interleave    & 0, 2, 4                         & --             & 0       \\
    Surrogate            & RF, GP                          & --             & GP      \\
    Extratrees           & TRUE, FALSE                     & Surrogate = RF &  FALSE  \\
    Trees                & 10, 500                         & Surrogate = RF & 500     \\
    Variance estimator   & Jackknife, ESD, LTV             & Surrogate = RF &  Jack    \\
    Kernel               & \makecell[l]{Gauss, Matérn~$3/2$,                            \\ Matérn~$5/2$, Exp}   & Surrogate = GP & Gauss \\
    Nugget               & 0, 1e-3, 1e-8                   & Surrogate = GP & 0       \\
    Scaling              & TRUE, FALSE                     & Surrogate = GP & FALSE   \\
    Acquisition function & EI, LCB, PI, mean               & --             & EI    \\
    $\lambda$            & 1, 3, 10                        & Acqf = LCB     & 1     \\
    Optimizer            & \makecell[l]{RS 1000, RS, LS,                              \\DIRECT, CMA-ES, L-BFGS-B} & -- & RS 1000 \\
    $\epsilon$ decay     & TRUE, FALSE                     & Acqf = EI      &  FALSE      \\
    $\lambda$ decay      & TRUE, FALSE                     & Acqf = LCB     &  FALSE     \\
    \bottomrule
  \end{tabular}
\end{table}

%% file: supplements/mixed_search_space.tex
\begin{table}[H]
  \centering
  \caption{Search space for mixed-hierarchical instances.}
  \label{tab:mlr3mbo-search-space-mixed}
  \begin{tabular}{lllll}
    \toprule
    Parameter            & Values                 & Dependency & Default \\
    \midrule
    Input trafo          & None, unitcube         & --         & None    \\
    Output trafo         & None, standardize, log & --         & None    \\
    Init                 & Random, LHS, Sobol     & --         & Random  \\
    Init size            & 0.05, 0.10, 0.25       & --         & 0.25    \\
    Random interleave    & 0, 2, 4                & --         & 0       \\
    Variance estimator   & Jackknife, ESD, LTV    & --         & ESD     \\
    Trees                & 10, 500                & --         & 500     \\
    Acquisition function & EI, LCB, PI, mean      & --         & EI      \\
    $\lambda$            & 1, 3, 10               & Acqf = LCB & 1      \\
    Optimizer            & RS 1000, RS, FS, LS    & --         & RS 1000 \\
    $\epsilon$ decay     & TRUE, FALSE            & Acqf = EI  & FALSE   \\
    $\lambda$ decay      & TRUE, FALSE            & Acqf = LCB & FALSE      \\
    \bottomrule
  \end{tabular}
\end{table}

%% file: supplements/numeric_instances.tex
\begin{table}[H]
  \centering
  \caption{Summary of YAHPO-SO v1 numeric instances with data set name and search space dimension.}
  \label{tab:numeric_instances}
  \begin{tabular}{lrlr}
    \toprule
    Scenario      & Instance & Data set      & Dimension \\
    \midrule
    lcbench       & 189873   & dionis        & 7         \\
    lcbench       & 189906   & segment       & 7         \\
    lcbench       & 167168   & vehicle       & 7         \\
    rbv2\_rpart   & 14       & mfeat-fourier & 5         \\
    rbv2\_rpart   & 40499    & texture       & 5         \\
    rbv2\_xgboost & 12       & mfeat-factors & 14        \\
    rbv2\_xgboost & 1501     & semeion       & 14        \\
    rbv2\_xgboost & 40499    & texture       & 14        \\
    \bottomrule
  \end{tabular}
\end{table}

%% file: supplements/mixed_instances.tex
\begin{table}[H]
  \centering
  \caption{Summary of YAHPO-SO v1 mixed-hierarchical instances with data set name and search space dimension.}
  \label{tab:mixed_instances}
  \begin{tabular}{lrlr}
    \toprule
    Scenario      & Instance & Data set                & Dimension \\
    \midrule
    lcbench       & 189873   & dionis                  & 7         \\
    lcbench       & 189906   & segment                 & 7         \\
    lcbench       & 167168   & vehicle                 & 7         \\
    nb301         & CIFAR10  & CIFAR10                 & 34        \\
    rbv2\_ranger  & 16       & mfeat-karhunen          & 8         \\
    rbv2\_ranger  & 42       & soybean                 & 8         \\
    rbv2\_rpart   & 14       & mfeat-fourier           & 5         \\
    rbv2\_rpart   & 40499    & texture                 & 5         \\
    rbv2\_super   & 1457     & amazon-commerce-reviews & 38        \\
    rbv2\_super   & 15       & breast-w                & 38        \\
    rbv2\_super   & 1063     & kc2                     & 38        \\
    rbv2\_xgboost & 12       & mfeat-factors           & 14        \\
    rbv2\_xgboost & 16       & mfeat-karhunen          & 14        \\
    rbv2\_xgboost & 1501     & semeion                 & 14        \\
    \bottomrule
  \end{tabular}
\end{table}

%% file: supplements/numeric_ablation.tex
\begin{table}[H]
  \centering
  \caption{Ablation study on the numeric search space.
    The `$\Delta$ start' column shows the change in RSNS when one parameter of the start configuration is changed.
    The `$\Delta$ final' column shows the change in RSNS when one parameter of the final configuration is changed.
    Dependent parameters with unsatified dependencies in the start or final configuration are omitted.
    When a parent parameters was changed, the dependent parameters were only changed to their default values e.g. GP to RF with 500 trees and jackknife variance estimator.
  }
  \label{tab:ablation-numeric}
  \begin{tabular}{llrr}
    \toprule
    Parameter            & Value       & $\Delta$ start & $\Delta$ final \\
    \midrule
    Input trafo          & Unitcube    & 0.00           & -0.02          \\
    Output trafo         & Log         & 0.06           & --             \\
                         & None        & --             & -0.21          \\
                         & Standardize & 0.00           & -0.20          \\
    Init                 & LHS         & 0.01           & -0.02          \\
                         & Sobol       & 0.00           & -0.01          \\
    Init size            & 0.05        & 0.05           & --             \\
                         & 0.10        & 0.05           & -0.03          \\
                         & 0.25        & --             & -0.01          \\
    Random interleave    & 2           & -0.17          & -0.04          \\
                         & 4           & -0.07          & -0.02          \\
    Surrogate            & RF          & -0.18          & -0.20          \\
    Kernel               & Exp         & -0.07          & -0.12          \\
                         & Gauss       & --             & -0.05          \\
                         & Matérn~$3/2$  & 0.06           & --             \\
                         & Matérn~$5/2$  & 0.05           & -0.02          \\
    Nugget               & 0           & --             & -0.05          \\
                         & $10^{-3}$        & 0.01           & -0.05          \\
                         & $10^{-8}$        & 0.00           & --             \\
    Scaling              & TRUE        & 0.01           & -0.22          \\
    Acquisition function & CB          & 0.01           & --             \\
                         & EI          & --             & -0.09          \\
                         & Mean        & -0.02          & -0.32          \\
                         & PI          & -0.01          & -1.10          \\
    $\lambda$            & 1           & --             & -0.16          \\
                         & 10          & --             & -0.13          \\
    $\epsilon$ decay     & TRUE        & -0.08          & --             \\
    $\lambda$ decay      & TRUE        & --             & -0.12          \\
    Optimizer            & CMA-ES      & 0.24           & --             \\
                         & DIRECT      & 0.10           & -0.11          \\
                         & L-BFGS-B    & 0.18           & -0.59          \\
                         & LS          & 0.12           & -0.01          \\
                         & RS          & 0.02           & -0.20          \\
                         & RS 1000     & --             & -0.29          \\
    \bottomrule
  \end{tabular}
\end{table}

%% file: supplements/mixed_ablation.tex
\begin{table}[H]
  \centering
  \caption{Ablation study on the mixed-hierarchical search space.
    The `$\Delta$ start' column shows the change in RSNS when one parameter of the start configuration is changed.
    The `$\Delta$ final' column shows the change in RSNS when one parameter of the final configuration is changed.
    Dependent parameters with unsatified dependencies in the start or final configuration are omitted.
    When a parent parameters was changed, the dependent parameters were only changed to their default values e.g. EI to LCB with a $\lambda$ value of 1 and no $\lambda$ decay.
  }
  \label{tab:ablation-mixed}
  \begin{tabular}{llrr}
    \toprule
    Parameter            & Value       & $\Delta$ start & $\Delta$ final \\
    \midrule
    Input trafo          & Unitcube    & -0.01          & -0.02          \\
    Output trafo         & Log         & 0.16           & --             \\
                         & None        & --             & -0.22          \\
                         & Standardize & 0.02           & -0.08          \\
    Init                 & LHS         & -0.01          & -0.04          \\
                         & Sobol       & 0.00           & -0.06          \\
    Init size            & 0.05        & 0.04           & --             \\
                         & 0.10        & 0.04           & -0.04          \\
                         & 0.25        & --             & -0.05          \\
    Random interleave    & 2           & -0.07          & -0.08          \\
                         & 4           & -0.01          & -0.03          \\
    Trees                & 10          & -0.07          & -0.15          \\
    Variance estimator   & ESD         & --             & -0.05          \\
                         & Jack        & 0.01           & -0.08          \\
                         & LTV         & 0.01           & --             \\
    Acquisition function & CB          & 0.10           & --             \\
                         & EI          & --             & -0.04          \\
                         & Mean        & 0.04           & -0.12          \\
                         & PI          & 0.09           & -0.07          \\
    $\lambda$            & 10          & --             & -0.02          \\
                         & 3           & --             & -0.05          \\
    $\epsilon$ decay     & TRUE        & -0.22          & --             \\
    $\lambda$ decay      & TRUE        & --             & -0.08          \\
    Optimizer            & LS          & 0.10           & --             \\
                         & RS          & 0.04           & -0.22          \\
                         & RS 1000     & --             & -0.27          \\
    \bottomrule
  \end{tabular}
\end{table}

%% file: supplements/software-versions.tex
\begin{table}[H]
  \centering
  \caption{Software packages referenced in this paper with their latest release versions.}
  \label{tab:software-versions}
  \begin{tabular}{llll}
    \toprule
    Package & Version & Language & Source \\
    \midrule
    mlrMBO & 1.1.6 & R & CRAN \\
    mlr3 & 1.6.0 & R & CRAN \\
    mlr3tuning & 1.6.0 & R & CRAN \\
    mlr3pipelines & 0.11.0 & R & CRAN \\
    mlr3learners & 0.14.0 & R & CRAN \\
    mlr3verse & 0.3.1 & R & CRAN \\
    bbotk & 1.10.0 & R & CRAN \\
    paradox & 1.0.1 & R & CRAN \\
    ranger & 0.18.0 & R & CRAN \\
    nloptr & 2.2.1 & R & CRAN \\
    data.table & 1.18.2.1 & R & CRAN \\
    future & 1.70.0 & R & CRAN \\
    mirai & 2.6.1 & R & CRAN \\
    rush & 1.1.0 & R & CRAN \\
    DiceOptim & 2.1.2 & R & CRAN \\
    GPareto & 1.1.9 & R & CRAN \\
    rBayesianOptimization & 1.2.2 & R & CRAN \\
    ParBayesianOptimization & 1.2.6 & R & CRAN (archived) \\
    tune & 2.1.0 & R & CRAN \\
    tidyverse & 2.0.0 & R & CRAN \\
    SMAC3 & 2.4.0 & Python & PyPI \\
    HEBO & 0.3.6 & Python & PyPI \\
    Ax & 1.2.4 & Python & PyPI \\
    BoTorch & 0.17.2 & Python & PyPI \\
    Optuna & 4.8.0 & Python & PyPI \\
    Hyperopt & 0.2.7 & Python & PyPI \\
    Dragonfly & 0.1.7 & Python & PyPI \\
    NUBO & 1.2.1 & Python & PyPI \\
    BoFire & 0.3.1 & Python & PyPI \\
    BayBE & 0.14.3 & Python & PyPI \\
    BOA & 0.11.0 & Python & PyPI \\
    BayesOpt & 0.8 & C++ & GitHub \\
    BayesianOptimization.jl & 0.2.5 & Julia & GitHub \\
    bayesopt & R2025a & MATLAB & Built-in toolbox \\
    \bottomrule
  \end{tabular}
\end{table}

%% file: supplements/uncertainty_estimation.tex
\subsection*{Uncertainty Estimation}
\label{sec:uncertainty-estimation}

To estimate the uncertainty of a random forest prediction, three methods are available in \texttt{mlr3mbo}: Jackknife-after-bootstrap, ensemble variance, and the law of total variance.
Let $t_b$ be the trees in the forest, with $b \in \{1, \ldots, B\}$. 
Define the average of all predictors, i.e.\ the prediction of the random forest, to be $\bar{f}(x) = \frac{1}{B} \sum_{b=1}^B t_b(x)$. 

For \textbf{jackknife-after-bootstrap}\cite{wager2014}:
Let $N_{b,i}$ be the number of times observation $i$ occurs in-bag for tree $b$ in the bootstrap.
Then 
\begin{displaymath}
f_{-i}(x) = \frac{1}{B_i} \sum_{b: N_{b,i} = 0} t_b(x),  \qquad B_i = \left|\{b: N_{b,i} = 0\}\right|
\end{displaymath}
is the "leave-$i$-out" predictor using all trees for which observation $i$ was out-of-bag.
The predictive variance $\widehat{\sigma}^2_J(x)$ is then estimated by:

\begin{displaymath}
\widehat{\sigma}^2_J(x) = \max\left(\frac{n - 1}{n} \sum_{i=1}^{n} \left( f_{-i}(x) -  \bar{f}(x) \right)^2 - (e - 1)\frac{n}{B^2}\sum_{b=1}^{B}\left(t_b(x)-\bar{f}(x)\right)^2, 0\right)\textrm{,}
\end{displaymath}
where the second sum term is a finite sample Monte Carlo bias correction.

For \textbf{ensemble variance}\cite{hutter2014}:
Consider a fixed test observation $x$.
For each tree $t_b$, compute the empirical mean $\mu_b$ of the training data points that fall into the predicted leaf node of $x$.
This means in particular that $\mu_b=t_b(x)$ and $\bar{f}(x)=\frac{1}{B}\sum_{b=1}^{B}\mu_b$.
The (naive) ensemble variance is then
\begin{displaymath}
  \sigma_{\mathrm{ev}}^2(x)=\frac{1}{B-1}\sum_{b=1}^{B}\left(\mu_b-\bar{f}(x)\right)^2
\end{displaymath}

For \textbf{law of total variance}\cite{hutter2014}: Again, in a similar fashion, consider a fixed $x$.
For each tree $t_b$, compute the empirical mean $\mu_b$ and variance $\sigma_b^2$ of the training data points that fall into the predicted leaf node of $x$.
The variance prediction $\hat{\sigma}_{\mathrm{ltv}}^2$ is the mean of the variance $\sigma_b^2$ of the leaf nodes plus the variance of the means $\mu_b$ of the leaf nodes:

\begin{align*}
\hat{\sigma}_{\mathrm{ltv}}^2(x) &=\frac{1}{B} \sum_{b=1}^{B} \sigma_b^2
+ \frac{1}{B}\sum_{b=1}^{B}\left(\mu_b-\bar{f}(x)\right)^2\\
&=\frac{1}{B} \sum_{b=1}^{B} \left(\sigma_b^2 + \mu_b^2\right)-\bar{f}(x)^2
\end{align*}

This captures the noise each tree sees within its leaf and the disagreements between the trees.
To avoid underestimating uncertainty, sometimes a minimum variance, of e.g., $10^{-2}$ is assigned to leaf nodes with small $\sigma_b^2$.
This approach ensures that predictions in regions with limited data reflect the inherent uncertainty.

%% file: supplements/loop_function.tex
\subsection*{Putting It All Together}
\label{sec:loop_function}

As an example, we show how to build a simple EGO loop function that uses the provided building blocks.

\begin{minted}{r}
my_simple_ego = function(
    instance,
    surrogate,
    acq_function,
    acq_optimizer,
    init_design_size
  ) {

  # setting up the building blocks
  surrogate$archive = instance$archive # archive
  acq_function$surrogate = surrogate # surrogate model
  acq_optimizer$acq_function = acq_function # acquisition function

  search_space = instance$search_space

  # initial design
  design = generate_design_sobol(
    search_space, n = init_design_size)$data
  instance$eval_batch(design)

  # BO loop
  repeat {
    candidate = tryCatch({
      # update the surrogate model
      acq_function$surrogate$update()
      # update the acquisition function
      acq_function$update()
      # optimize the acquisition function to yield a new candidate
      acq_optimizer$optimize()
    }, mbo_error = function(mbo_error_condition) {
      generate_design_random(search_space, n = 1L)$data
    })

    # evaluate the candidate and add it to the archive
    tryCatch({
      instance$eval_batch(candidate)
    }, terminated_error = function(cond) {
      # eval_batch() throws a terminated_error if the instance is
      # already terminated, e.g. because of timeout
    })
    if (instance$is_terminated) break
  }

  return(instance)
}
\end{minted}

%% file: supplements/output_transformations.tex
\subsection*{Output Transformations}
\label{sec:output-transformations}

Transformations of the target values are a common technique to improve the performance of the surrogate model~\cite{jones1998}.
When the surrogate model is trained on the transformed target values, one would typically apply the inverse transformation to the predictive mean and standard deviation before passing them to the acquisition function, to get values on the original scale.
Alternatively, predictions can be kept on the transformed scale and the acquisition function can be defined directly in that space.
LogEI, for example, is explicitly designed to operate on log-transformed targets.

\paragraph{Log-Transformed Surrogate Models}

In many hyperparameter optimization scenarios, the distribution of the target variable (e.g., runtime or loss) is heavy right-tailed due to some very poorly performing configurations.
These configurations can correspond to timeouts or models that fail to train and are typically assigned very large loss or runtime values.
Log-transforming the target values compresses these extremes, makes the marginal distribution more symmetric and Gaussian-like, and can reduce heteroscedasticity.
However, since target values are required to be positive for log transformation, it may be possible to add an offset when negative objective values can be expected.
\texttt{mlr3mbo} therefore scales output values to $(0, 1)$ before applying the log transformation.

GP surrogate models tend to be more accurate in this setting when they are trained on $\log(y)$ instead of $y$~\cite{jones1998,hutter2009}, because the GP assumptions of Gaussian, homoscedastic residuals are better satisfied.
Similarly, random forests tend to find more balanced splits on log-transformed targets and can perform better on heavy-tailed objectives~\cite{lindauer2019}.

When LCB or PI are computed directly on the log-transformed surrogate, no inverse transformation of the predictions is required and the acquisition function is simply interpreted as operating in log-space.
If, instead, these acquisition functions are to be applied on the original scale $y$, the predictive distribution in log-space must be mapped back to $y$ (e.g., using the log-normal moment formulas).

\paragraph{LogEI Acquisition Function}

The LogEI acquisition function is the expected improvement on the original objective scale when the surrogate is trained on log-transformed target values~\cite{hutter2009}. This means it is
\begin{displaymath}
\alpha_{\mathrm{LogEI}}(\xv) = \mathbb{E} \left[ \max \left( f_{\min} - Y(\xv), 0 \right) \right],
\end{displaymath}
with $Y(\xv)$ interpreted as a log-normal distribution and $\ln Y(\xv) \sim \mathcal{N}(\mu(\xv), \sigma^{2}(\xv))$.
The LogEI acquisition function is then defined as
\begin{equation}
  \alpha_{\mathrm{LogEI}}(\xv)
  = f_{\min}\,\Phi\big(v(\xv)\big)
    - \exp\!\Big(\mu(\xv) + \tfrac{1}{2}\sigma^{2}(\xv)\Big)\,
      \Phi\big(v(\xv) - \sigma(\xv)\big),
\end{equation}
where
\begin{equation}
  v(\xv) = \frac{\ln f_{\min} - \mu(\xv)}{\sigma(\xv)}.
\end{equation}
Here, $\Phi$ denotes the cumulative distribution function of the standard normal distribution, and $f_{\min}$ is the best function value in the archive.
Thus, LogEI computes the expected improvement on the original scale under a log-normal model, which can better capture heavy right tails than a surrogate model defined directly on $y$.
Both the original SMAC and SMAC3 (in its SMAC4HPO facade) use LogEI as the default acquisition function whenever a log-transform of the performance values is employed~\cite{hutter2011,lindauer2022}. LogEI should not be confused with the logarithm of the EI, which is also used by various methods~\cite{watanabe2024}.

%% file: supplements/local_search.tex
\subsection*{Local Search}
\label{sec:local-search}

The local search (LS) acquisition function optimizer in the \texttt{bbotk} package works similarly to the LS implementation in SMAC\cite{lindauer2022} and can be used to optimize acquisition functions in mixed spaces.
To reduce function call overhead, $k$ LS runs are generated concurrently, so that as many points as possible can be passed to the SM in a single batch.

Each LS run is executed for a specified number of iterations.
In each iteration, the LS generates multiple candidate neighbors by mutating the current point / incumbent.
A neighbor is obtained by mutating exactly one parameter from the current point.
For numeric parameters, the parameter value is first scaled to $[0, 1]$, zero-mean Gaussian noise (with standard deviation governed by a control parameter) is added, and the result is then rescaled to the original domain.
Afterwards, the value is clipped to the corresponding lower and upper bounds.
For integer parameters, the same procedure is applied, but the result is rounded to the nearest integer.
For factor parameters, a new level is sampled uniformly from the levels that differ from the current one.
For logical parameters, the value is simply flipped.

Hierarchical dependencies in the search space are handled as follows.
Only active parameters are eligible for mutation.
After a mutation occurs, all conditions in the search space are checked in topological order.
If a condition is not satisfied, the corresponding parameter is marked as inactive.
Conversely, if all conditions for a parameter are satisfied but it is marked as inactive, it is assigned a random valid value.

After the neighbors are generated, they are evaluated using the acquisition function.
The number of candidates passed to the SM simultaneously is therefore $k$ times the number of neighbors generated per LS run. 
For each LS, the best neighbor is selected as the new incumbent unless all neighbors are worse than the current incumbent, in which case the incumbent is retained.
To avoid getting stuck in local optima, the algorithm employs a restart mechanism: 
For each LS, the number of consecutive iterations without improvement is tracked, and if this 
exceeds a specified threshold, the search is restarted from a new random point in the search space.

%% file: supplements/technical_details.tex
\subsection*{Technical Details}
\label{sec:technical-details}

This section provides more details on the technical settings that were not considered in the CD procedure.
The function evaluation budget of the acquisition function optimization was set to $100 \cdot d^2$ evaluations.
DIRECT and L-BFGS-B were randomly restarted for $5 \cdot d$ times and CMA-ES and LS were restarted until the global budget was exhausted.
The random forest implementation in mlr3mbo uses the ranger package\cite{wright2017} with the following settings different from the default: \texttt{sample.fraction = 1}, \texttt{min.node.size = 3}, \texttt{min.bucket = 3}, \texttt{mtry.ratio = 5/6}.
The Gaussian process implementation in mlr3mbo uses the DiceKriging package\cite{roustant2012} with the \texttt{optim.method} changed to \texttt{"gen"}.
The $\lambda$ and $\epsilon$ decays reduce the value of the parameters by 1\% in each iteration.